\documentclass{article}

\usepackage{PRIMEarxiv}

\usepackage[utf8]{inputenc} % allow utf-8 input
\usepackage[T1]{fontenc}    % use 8-bit T1 fonts
\usepackage{hyperref}       % hyperlinks
\usepackage{url}            % simple URL typesetting
\usepackage{booktabs}       % professional-quality tables
\usepackage{amsfonts}       % blackboard math symbols
\usepackage{nicefrac}       % compact symbols for 1/2, etc.
\usepackage{microtype}      % microtypography
\usepackage{lipsum}
\usepackage{fancyhdr}       % header
\usepackage{graphicx}       % graphics
\graphicspath{{media/}}     % organize your images and other figures under media/ folder
\usepackage{amsmath,amssymb}
\usepackage{algorithm}
\usepackage{algorithmic}
\usepackage{float}

\title{Language as a latent sequence: deep latent variable models for semi-supervised paraphrase generation
\thanks{\textit{\underline{Accepted by AI Open journal: \url{https://www.sciencedirect.com/science/article/pii/S2666651023000025}}}} 
}

\author{
  Jialin Yu, Alexandra I. Cristea, Anoushka Harit, Zhongtian Sun, \\ \textbf{Olanrewaju Tahir Aduragba, Lei Shi, Noura Al Moubayed} \\
  Department of Computer Science \\
  Durham University \\
  Durham, UK\\
}

\begin{document}
\maketitle

\begin{abstract}
% \lipsum[1]
% This paper explores deep latent variable models for semi-supervised paraphrase generation, where the missing target pair is modelled as a latent paraphrase sequence. We present a novel unsupervised model named variational sequence auto-encoding reconstruction (\textbf{VSAR}), which performs latent sequence inference given an observed text. To leverage information from text pairs, we introduce a supervised model named dual directional learning (\textbf{DDL}). Combining VSAR with DDL (\textbf{DDL+VSAR}) enables us to conduct semi-supervised learning; however, the combined model suffers from a cold-start problem. To combat this issue, we propose to deal with better weight initialisation, leading to a two-stage training scheme named knowledge reinforced training. Our empirical evaluations suggest that the combined model yields competitive performance against the state-of-the-art supervised baselines on complete data. Furthermore, in scenarios where only a fraction of the labelled pairs are available, our combined model consistently outperforms the strong supervised model baseline (\textbf{DDL}) by a significant margin.
This paper explores deep latent variable models for semi-supervised paraphrase generation, where the missing target pair {for unlabelled data} is modelled as a latent paraphrase sequence. We present a novel unsupervised model named \textit{variational sequence auto-encoding reconstruction} (\textbf{VSAR}), which performs latent sequence inference given an observed text. To leverage information from text pairs, we {additionally} introduce a {novel} supervised model {we call \textit{dual directional learning}} (\textbf{DDL}){, which is designed to integrate with our proposed VSAR model}. Combining VSAR with DDL (\textbf{DDL+VSAR}) enables us to conduct semi-supervised learning{. Still}, the combined model suffers from a cold-start problem. To {further} combat this issue, we propose {an improved} weight initialisation {solution}, leading to a {novel} two-stage training scheme {we call} \textit{knowledge-reinforced-learning} {(\textbf{KRL})}. Our empirical evaluations suggest that the combined model yields competitive performance against the state-of-the-art supervised baselines on complete data. Furthermore, in scenarios where only a fraction of the labelled pairs are available, our combined model consistently outperforms the strong supervised model baseline (\textbf{DDL}) by a significant margin ($p <.05$; Wilcoxon test). Our code is publicly available at \url{https://github.com/jialin-yu/latent-sequence-paraphrase}.
\end{abstract}

% keywords can be removed

\section{Introduction}

Paraphrase generation is an important Natural Language Processing (NLP) problem, useful in many NLP applications, such as question answering \cite{dong2017learning}, information retrieval \cite{lee2006beyond}, information extraction \cite{yao2014information} and summarisation \cite{liu2008understanding}. Natural language itself is complicated and may be expressed in various alternative surface forms of the same underlying semantic content \cite{miller2019explanation,hosking2022hierarchical}. Hence {it} is critically important to integrate the paraphrase generation model as a component in real-world NLP systems{,} to offer robust responses to end users' inputs. Traditional solutions to paraphrase generation are generally rule-based \cite{kauchak2006paraphrasing,narayan2016paraphrase}, utilising lexical resources, such as WordNet \cite{miller1995wordnet}, to find word replacements. The recent trend brings to fore neural network models \cite{kumar2020syntax,zhou2021paraphrase,meng2021conrpg,su2021keep}, which are typically based on a sequence-to-sequence learning paradigm \cite{sutskever2014sequence}.

These models have achieved remarkable success for paraphrase generation, due to complex architectures and sophisticated conditioning mechanisms, \textit{e.g. soft, hard and self-attention}. However, the advancement of such models is primarily based on the availability of large-scale labelled data pairs. Instead, \textit{this paper explores semi-supervised learning scenarios, where only a fraction of the labels are available}. This semi-supervised learning setting is favourable and extremely useful for industry scenarios \cite{zhu2005semi,van2020survey}, due to the effort {in terms of} time and money to obtain good quality human annotations. A semi-supervised learning model often consists of two components: an unsupervised learning model and a supervised learning model. 

Thus, for the unsupervised learning part, we propose a novel deep generative model, motivated by the classic variational autoencoder (VAE) \cite{kingma2013auto,rezende2014stochastic,mnih2014neural}, with an additional structural assumption tailored towards modelling language sequences, named \textit{variational sequence auto-encoding reconstruction} (\textbf{VSAR}). To further explain, traditional VAEs typically embed data representations in a fixed latent space, with the general purpose of dimensionality reduction \cite{hinton2006reducing}. {Here we consider instead} a latent variable in the form of a discrete language sequence with various lengths. This assumption enforces more structural information to be adopted in the model training process and can additionally enhance the model interpretability, as language is naturally preserved as discrete variables \cite{fu2020paraphrase}. Following the recent prior works successfully incorporating discrete latent variables to improve paraphrasing \cite{hosking2021factorising, hosking2022hierarchical}, we propose our model, the VSAR model, aimed to contain a more expressive form of discrete latent variable, shown in Figure \ref{fig: VSAR_Model}.

Furthermore, for the supervised learning part, motivated by dual learning \cite{he2016dual,su2019dual,su2020dual,su2020towards}, we propose a novel supervised model, named \textit{dual directional learning} (\textbf{DDL}) that can be integrated with our proposed VAE model, which shares a part of the learning loop of the VSAR model. Combining both unsupervised and supervised models enables semi-supervised learning, by exploiting VAE's ability to marginalise latent variables for unlabelled data.

Our main original contributions in this paper thus include:

\begin{itemize}
  \item presenting \textit{the first study on semi-supervised learning for paraphrasing with deep {discrete} latent variable models};
  \item introducing \textit{two novel models: VSAR (unsupervised) and DDL (supervised)}, which can be combined for semi-supervised learning;
  \item proposing \textit{a novel training scheme, knowledge-reinforced-learning (KRL)} to deal with {the} cold start problem in the combined {semi-supervised} model (DDL+VSAR);
  \item studying semi-supervised learning scenarios with the combined model on the full data and empirically showing that our model achieves competitive state-of-the-art results;
  \item presenting a study of semi-supervised scenarios {on} a fraction of the labelled data{(i.e., when incorporating unlabelled data), demonstrating} significantly better results {for our models} than {for} very strong supervised baselines.
\end{itemize}

\section{Related Work}
\label{sec: related-work}

\subsection{Paraphrase Generation}

Paraphrases express the surface forms of the underlying semantic content \cite{hosking2022hierarchical} and capture the essence of language diversity \cite{pavlick2015ppdb}. Early work on automatic generation of paraphrase{s} are generally rule-based \cite{kauchak2006paraphrasing,narayan2016paraphrase}, but the recent trend brings to {the} fore neural network solutions \cite{gupta2018deep,fu2020paraphrase,kumar2020syntax,meng2021conrpg,su2021keep, hosking2021factorising, hosking2022hierarchical, chen2022mcpg, xie2023visual}. Current research for paraphrasing mainly focuses on supervised methods, which require the availability of a large number of source and target pairs. In this work, we {instead} explore a semi-supervised paraphrasing method, where only a fraction of source and target pairs are observed, and where a large number of  unlabelled source text{s} exist. We made an assumption that each missing target text can be considered as a latent variable in deep generative models. {Thus, for unsupervised data, each missing paraphrase output is modelled as a latent variable. Compared to the standard approach, where the semantics of a sentence is presented as a dense high-dimensional vector, this assumption enforces the model to learn more \textit{structured} representations. Hence, our proposed model can be considered as a type of deep latent structure model}\cite{martins-etal-2019-latent}. In this paper, we present two models and combine them for paraphrasing: one for unsupervised learning and one for supervised learning. Our combined model extends {the idea in} \cite{miao2016language, fu2020paraphrase}; {compared with} \cite{fu2020paraphrase}, {our model utilises conversely a more natural language structure (an ordered sequence other than an unordered bag of words); compared with} \cite{miao2016language}, {our model utilises a self-attention mechanism other than convolution operations and has the benefit of being able to model various lengths of latent sequences, rather than a fixed length.}

%and models jointly the distribution of source and target, instead of the conditional probability of a target, given the source. 
Furthermore, our combined model is associated with prior works that introduce a discrete latent variable \cite{hosking2021factorising,hosking2022hierarchical}, and it uses an arguably more expressive latent variable, in the form of language {outputs}.

\subsection{Deep Latent Variable Models for Text}

Deep latent variable models have been studied for text modelling \cite{miao2016neural,kim2018tutorial}. The most common and widely adopted latent variable model is the standard VAE model with a Gaussian prior \cite{bowman2015generating}, which suffers from posterior collapse \cite{dieng2019avoiding,he2019lagging}. Multiple studies have been conducted to combat this issue \cite{higgins2016beta,razavi2019preventing,wang2021posterior}. In particular, $\beta$-VAE \cite{higgins2016beta} introduces a penalty term to balance VAE reconstruction and prior  regularisation intuitively  and is adopted as one of our baselines.

While much of the research focuses on continuous latent variable models \cite{miao2016neural,kim2018tutorial}, the text is naturally presented in discrete form and may not be well represented with continuous latent variables. Early work on discrete deep latent variable models \cite{miao2016language,wen2017latent} adopted the REINFORCE algorithm \cite{mnih2014neural,mnih2014recurrent}; however, it suffers from very high variance. With the recent advancement in statistical relaxation techniques, {the} Gumbel {t}rick \cite{jang2016categorical,maddison2016concrete} was utilised, to model discrete structures in the latent variable model of the text \cite{choi2018learning,fu2020paraphrase,he2020probabilistic,mercatali2021disentangling}. Our work adopts {the} Gumbel-{t}rick with subset sampling \cite{xie2019reparameterizable} for natural language generation tasks and, \textit{for the first time, studies {discrete language sequences as} a latent variable  for the paraphrasing task}. Our proposed model is strongly associated with \cite{miao2016language, he2020probabilistic}; however, we study the problem under the semi-supervised setup for the paraphrase generation tasks. Furthermore, we present a novel inference algorithm (our knowledge-reinforced-learning (KRL) scheme) to help aid learning in deep generative models and achieve competitive performance for both full data and {incomplete (fractional)} data settings. {In terms of deep latent variable models for text, there has been a recent surge of interest in learning discrete latent structures} \cite{niculae2023discrete}. {In this paper, we contribute thus to this research field by modelling the latent variable as a language sequence. Additionally, some recent work focused on combining latent variable models with diffusion models, to achieve state-of-the-art performance} \cite{huang2023make, schneider2023mo} {for text-to-audio generation tasks.} 

\begin{figure}[!h]
\centering
\resizebox{0.95\columnwidth}{!}{
\includegraphics{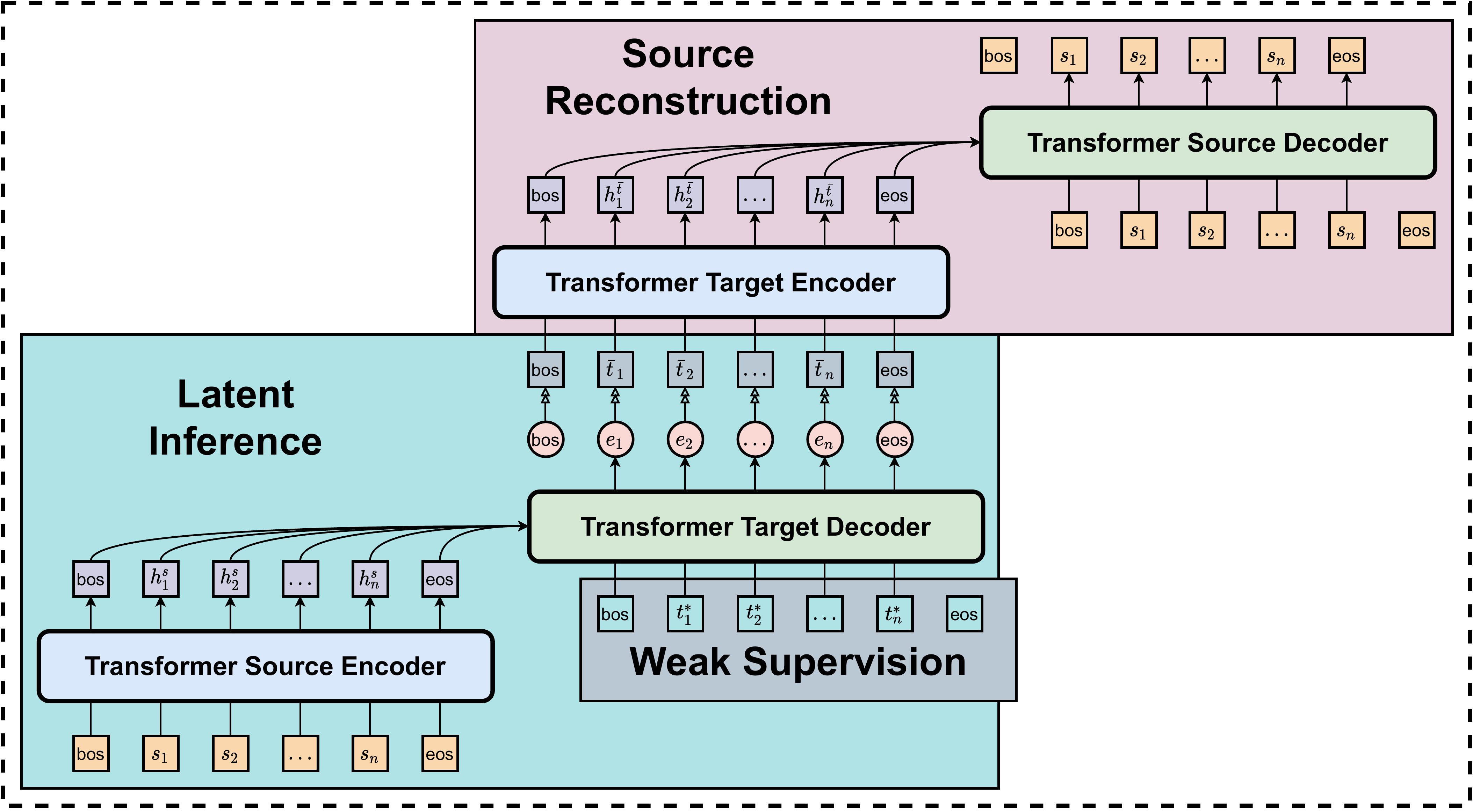}}
\caption{\bf Variational Sequence Auto-Encoding Reconstruction Model.}
\label{fig: VSAR_Model}
\end{figure}

\section{Variational Sequence Auto-Encoding Reconstruction (VSAR)}
\label{sec: VSAR}

In this section, we present the VSAR model (Figure \ref{fig: VSAR_Model})\footnote{The language model prior and weak supervision decoding is omitted, for clarity.}. The model consists of four separate neural network models - a source encoder, a target decoder, a target encoder, and a source decoder. Under the unsupervised learning setup, we only observe source text $\boldsymbol{s}$ and no target text $\boldsymbol{t}$. We reformulate the problem of modelling fully observed source text $\boldsymbol{s}$, as modelling {the} partially observed parallel source text $\boldsymbol{s}$ and its associated latent target pair $\bar{\boldsymbol{t}}$. We adopt Bayesian inference{,} to marginalise the latent target string $\bar{\boldsymbol{t}}$ from the joint probability distribution $p_{\boldsymbol{\theta}}(\boldsymbol{s}, \bar{\boldsymbol{t}})$, based on Equation \ref{equ: vasr_marginal_likelihood}, as shown in Figure \ref{fig: VSAR_Model}. 

Thus, in the VSAR model, the \textbf{latent inference} network, parameterised as $q_{\boldsymbol{\phi}}(\bar{\boldsymbol{t}} \vert \boldsymbol{s})$, takes source text $\boldsymbol{s}$ and generates a latent target sample $\bar{\boldsymbol{t}}$. The \textbf{source reconstruction} network, parameterised as $p_{\boldsymbol{\theta}}(\boldsymbol{s} \vert \bar{\boldsymbol{t}})$, reconstructs the observed source text $\boldsymbol{s}$ back, based on the latent target sample $\bar{\boldsymbol{t}}$. As the prior distribution, a language model is pre-trained on unlabelled source text corpus{,} to approximate the prior distribution $p(\bar{\boldsymbol{t}})$\footnote{We leverage linguistic knowledge of {the} paraphrase generation task, in which a paraphrase text string can be considered as its own paraphrase.}. The prior is introduced for regularisation purposes \cite{miao2016language,he2020probabilistic}, which enforces {that} samples are more likely to be {'}reasonable{'} natural language text.

Motivated by the benefits of \textbf{parameters sharing} in multi-task learning for natural language generation \cite{luong2015multi,guo2018dynamic,guo2018soft,wang2020multi}, we share model parameters for the source encoder and the target encoder, denoted as $\boldsymbol{f_{encode}}$; similarly, we share model parameters for the source decoder and the target decoder, denoted as $\boldsymbol{f_{decode}}$. In the following sections, we use $\boldsymbol{f_{encode}}$ and $\boldsymbol{f_{decode}}$ to represent all encoders and decoders in the VSAR model, respectively. 

\subsection{Weak Supervision}
\label{sec: weak_supervision}

In the VSAR model, we empirically found that the quality of {the} latent sequence $\bar{\boldsymbol{t}}$ is very unstable, especially at the beginning of the training. To combat this issue, motivated by the idea of weak supervision \cite{du2020self,chang2021jointly}, we propose to use pseudo-labels to guide VSAR throughout training. Before each model {performs} {the} forward{-}pass {using the back-propagation algorithm}, we first assign pseudo-labels to each token in {the} unobserved latent target sample $\bar{\boldsymbol{t}}$ {based on} the current model parameter {(from the previous iteration)}. The pseudo-labels are detached from the computational graph; hence no gradient is updated during the weak supervision process. The pseudo-labels can be considered as a weak supervision signal for `teacher forcing training' \cite{williams1989learning}.

The encoder model takes the source string $\boldsymbol{s} = (s_{1}, ..., s_{n})$ as input and produces its corresponding contextual vector $\boldsymbol{h^{s}} = (h_{1}^{s}, ..., h_{n}^{s})$:

\begin{equation}
\begin{aligned}
    \boldsymbol{h^{s}} &=\boldsymbol{f_{encode}}(\boldsymbol{s})
\end{aligned}
\label{equ: paraphrase_inference_encoder}
\end{equation}

We adopt a greedy decoding scheme to assign pseudo-target labels $\boldsymbol{t^{\ast}}$ and assume that a good paraphrase ought to have a similar length as the original sentence \cite{burrows2013paraphrase,cao2017joint}; such that $\boldsymbol{t^{\ast}} = (t_{1}^{\ast}, ..., t_{n}^{\ast})$. Let $t_{i}^{\ast}$ be the $i^{th}$ word in the pseudo target sequence; we construct this sequence in an auto-regressive manner:

\begin{equation}
\begin{aligned}
    t_{i}^{\ast} = \boldsymbol{f_{decode}}(\boldsymbol{h^{s}}; t_{1:i-1}^{\ast}) 
\end{aligned}
\label{equ: weak_supervision_decoder}
\end{equation}

\subsection{Target Inference}

Once the pseudo-target labels $\boldsymbol{t^{\ast}}$ are assigned, we perform latent variable inference with the latent inference network. Since the source string $\boldsymbol{s}$ remains the same{,} we reuse the value of the contextual vector $\boldsymbol{h^{s}}$ in the weak supervision section. Let $\bar{t}_{j}$ be the $j^{th}$ words in the latent sample and $e_{j}$ be the corresponding output of the target decoder model. We construct the latent sample $\bar{\boldsymbol{t}}$ using contextual vector $\boldsymbol{h_{s}}$ and all $t_{1:j-1}^{\ast}$ words in the pseudo-labels: 

\begin{equation}
\begin{aligned}
    e_{j} &= \boldsymbol{f_{decode}}(\boldsymbol{h_{s}}; t_{1:j-1}^{\ast}) \\
   \bar{t}_{j} &\sim \textit{Gumbel}\text{-}\text{TOP}k(e_{j}, \tau)
\end{aligned}
\label{equ: paraphrase_inference_decoder}
\end{equation}

{Here,} $\bar{t}_{i}$ is drawn via the Gumbel {t}rick \cite{jang2016categorical,maddison2016concrete} with temperature $\tau$ as an additional hyper-parameter, which controls the probability distribution of the samples. At a high temperature $\tau$, we equivalently sample from a uniform distribution; at a low temperature $\tau$, we equivalently sample from a categorical distribution. {Due to the enormous size of the vocabulary, sampling from this simplex can be difficult, hence we further adopt} the TOP-$k$ subset sampling technique \cite{xie2019reparameterizable} to improve sampling efficiency.

We explore two different schemes commonly used in the literature: (1) we use a fixed temperature $\tau$ of $0.1$, as in \cite{chen2018learning}; and (2) we gradually anneal the temperature $\tau$ from a high temperature of $10$ to a low temperature of $0.01$, as in \cite{balin2019concrete}. Our empirical results suggest that annealing the temperature $\tau$ during training yields significantly better results ($p <.05$; Wilcoxon test){, which} are thus used to report the final results. We use a $k$-value of $10${,} as suggested in \cite{fu2020paraphrase}.

\subsection{Source Reconstruction}

For the source reconstruction network, the encoder model takes the latent target sequence string $\bar{\boldsymbol{t}} = (\bar{t}_{1}, ..., \bar{t}_{n})$ as input and produces its corresponding contextual vector $\boldsymbol{h^{\bar{t}}} = (h_{1}^{\bar{t}}, ..., h_{n}^{\bar{t}})$:

\begin{equation}
\begin{aligned}
    \boldsymbol{h^{\bar{t}}} &= \boldsymbol{f_{encode}}(\bar{\boldsymbol{t}})
\end{aligned}
\label{equ: sequence_reconstruction_encoder}
\end{equation}

Let $\hat{s}_{k}$ be the $k^{th}$ word in the reconstructed source string, during the training; we retrieve the reconstructed source string $\hat{\boldsymbol{s}}$ via:

\begin{equation}
\begin{aligned}
    \hat{s}_{k} &= \boldsymbol{f_{decode}}(\boldsymbol{h^{\bar{t}}}; s_{1:k-1})
\end{aligned}
\label{equ: sequence_reconstruction_decoder}
\end{equation}

\subsection{Learning and Inference for VSAR}

In the SVAR model, there are two sets of parameters, $\boldsymbol{\phi}$ and $\boldsymbol{\theta}$, which are required to be updated. Let $\boldsymbol{S}$ be the observed random variable for the source text, $\boldsymbol{\bar{T}}$ be the latent random variable for the target text, and $N$ be the total number of the unlabelled source text. We have the following joint likelihood for the SVAR model, parameterised by $\boldsymbol{\theta}$:

\begin{equation}
\begin{aligned}
    p(\boldsymbol{S}, \boldsymbol{\bar{T}}; \boldsymbol{\theta}) = \prod_{i=1}^{N} p(s_{(i)} \vert \bar{t}_{(i)}; \boldsymbol{\theta}) p(\bar{t}_{(i)})
\end{aligned}
\label{equ: vasr_joint_likelihood}
\end{equation}

The log marginal likelihood $\boldsymbol{L_{1}}$ of the observed data that  will be approximated during training is $\log p(\boldsymbol{S}; \boldsymbol{\theta})$. We adopt amortised variational inference \cite{kingma2013auto,rezende2014stochastic,mnih2014neural} and build a surrogate function{,} approximated with a neural network $q(\boldsymbol{\bar{T}} \vert \boldsymbol{S}; \boldsymbol{\phi})$, parameterised by $\boldsymbol{\phi}$, to derive the evidence lower bound (ELBO) for the joint likelihood:

\begin{equation}
\begin{aligned}
    & \boldsymbol{L_{1}} = \log \sum_{\boldsymbol{\bar{T}}} p(\boldsymbol{S}, \boldsymbol{\bar{T}}; \boldsymbol{\theta}) \geq \mathcal{L}_{ELBO} (\boldsymbol{S}, \boldsymbol{\bar{T}}; \boldsymbol{\theta}, \boldsymbol{\phi}) \\ &= \sum_{i=1}^{N}\{\mathbb{E}_{q(\bar{t} \vert s_{(i)}; \boldsymbol{\phi})}[\log p(s_{(i)} \vert \bar{t}; \boldsymbol{\theta})] - \mathbb{D}_{KL}[q(\bar{t} \vert s_{(i)}; \boldsymbol{\phi}) \vert  \vert p(\bar{t})]\}
\end{aligned}
\label{equ: vasr_marginal_likelihood}
\end{equation}

The most common variational family in the VAE framework relies on the reparameterisation trick \cite{kingma2013auto}, which is not applicable {to} the non-differentiable discrete latent variable. An approach for optimising learning with such latent variables uses the REINFORCE algorithm \cite{mnih2014neural,mnih2014recurrent}; however, this algorithm generally suffers from high variance. In this paper, we instead use Gumbel-Softmax \cite{jang2016categorical,maddison2016concrete} with differentiable subset sampling \cite{xie2019reparameterizable}{,} to retrieve top-$k$ samples without replacement. Nevertheless, since sampling a one-hot form vector induces high variance, we apply the straight-through technique \cite{bengio2013estimating} as a biased estimator of the gradient, to combat this variance.

During training, while optimising the log-likelihood, we perform learning ($\boldsymbol{\theta}$) and inference ($\boldsymbol{\phi}$) at the same time. The parameters are jointly optimised with the same optimiser. Since we are sharing parameters in our model, in practice, we are updating the same set of parameters (shared by $\boldsymbol{\theta}$ and $\boldsymbol{\phi}$) with source data only.

\section{Dual Directional Learning (DDL)}
\label{sec: ddl}

In this section, we introduce the Dual Directional Learning (DDL) model, which we use for supervised paraphrase generation. The DDL model consists of two sets of standard Transformer models \cite{vaswani2017attention}, each with its own {two} separate neural networks - an encoder and a decoder. We perform standard sequence-to-sequence learning, with {the} fully observed parallel source text $\boldsymbol{s}$ and its associated target pair $\boldsymbol{t}$, in dual directions. The \textbf{target generation} network $p_{\boldsymbol{\theta_{t \vert s}}}(\boldsymbol{t} \vert \boldsymbol{s})$ takes source text $\boldsymbol{s}$ as input and generates target text $\boldsymbol{t}${;} and the \textbf{source generation} network $p_{\boldsymbol{\theta_{s \vert t}}}(\boldsymbol{s} \vert \boldsymbol{t})$ takes target text $\boldsymbol{t}$ as input and generates source text $\boldsymbol{s}$. 

\subsection{Parameter Learning}

In the DDL model, there are two sets of parameters, $\boldsymbol{\theta_{s \vert t}}$ and $\boldsymbol{\theta_{t \vert s}}$, which are required to be updated. Let $\boldsymbol{S}$ be the observed random variable for source text, $\boldsymbol{T}$ be the observed random variable for target text, and $M$ be the number of labelled pairs; we then have the following conditional likelihood for our DDL model:

\begin{equation}
\begin{aligned}
    & p(\boldsymbol{S}  \vert  \boldsymbol{T}; \boldsymbol{\theta_{s \vert t}}) = \prod_{i=1}^{M} p(s_{(i)} \vert t_{(i)}; \boldsymbol{\theta_{s \vert t}}) \\
    & p(\boldsymbol{T}  \vert  \boldsymbol{S}; \boldsymbol{\theta_{t \vert s}}) = \prod_{i=1}^{M} p(t_{(i)} \vert s_{(i)}; \boldsymbol{\theta_{t \vert s}})
\end{aligned}
\label{equ: ddl_joint_likelihood}
\end{equation}

The log conditional likelihood $\boldsymbol{L_{2}}$ of the observed data pairs can be jointly learnt during training as:

\begin{equation}
\begin{aligned}
    & \boldsymbol{L_{2}} = \sum_{i=1}^{M}(\log p(s_{(i)} \vert t_{(i)}; \boldsymbol{\theta_{s \vert t}}) + \log p(t_{(i)} \vert s_{(i)}; \boldsymbol{\theta_{t \vert s}}))
\end{aligned}
\label{equ: ddl_marginal_likelihood}
\end{equation}

During training, we perform dual learning ($\boldsymbol{\theta_{s \vert t}}$ and $\boldsymbol{\theta_{t \vert s}}$) at the same time and the parameters are jointly optimised with the same optimiser.

\subsection{Parameter Sharing}

Once again, motivated by the benefits of multi-task learning for natural language generation \cite{luong2015multi,guo2018dynamic,guo2018soft,wang2020multi}, we share model parameters for the target generation and the source generation network. Although sharing parameters is a very simple technique, as shown in Table \ref{tab: semi_quora} and Table \ref{tab: semi_mscoco}, the DDL model significantly improves the performance of paraphrase generation with respect to the Transformer baseline ($p <.05$; Wilcoxon test), which only handles sequence{-}to{-}sequence learning in a single direction.

\begin{figure*}[h!]
\centering
\resizebox{0.90\columnwidth}{!}{
\includegraphics{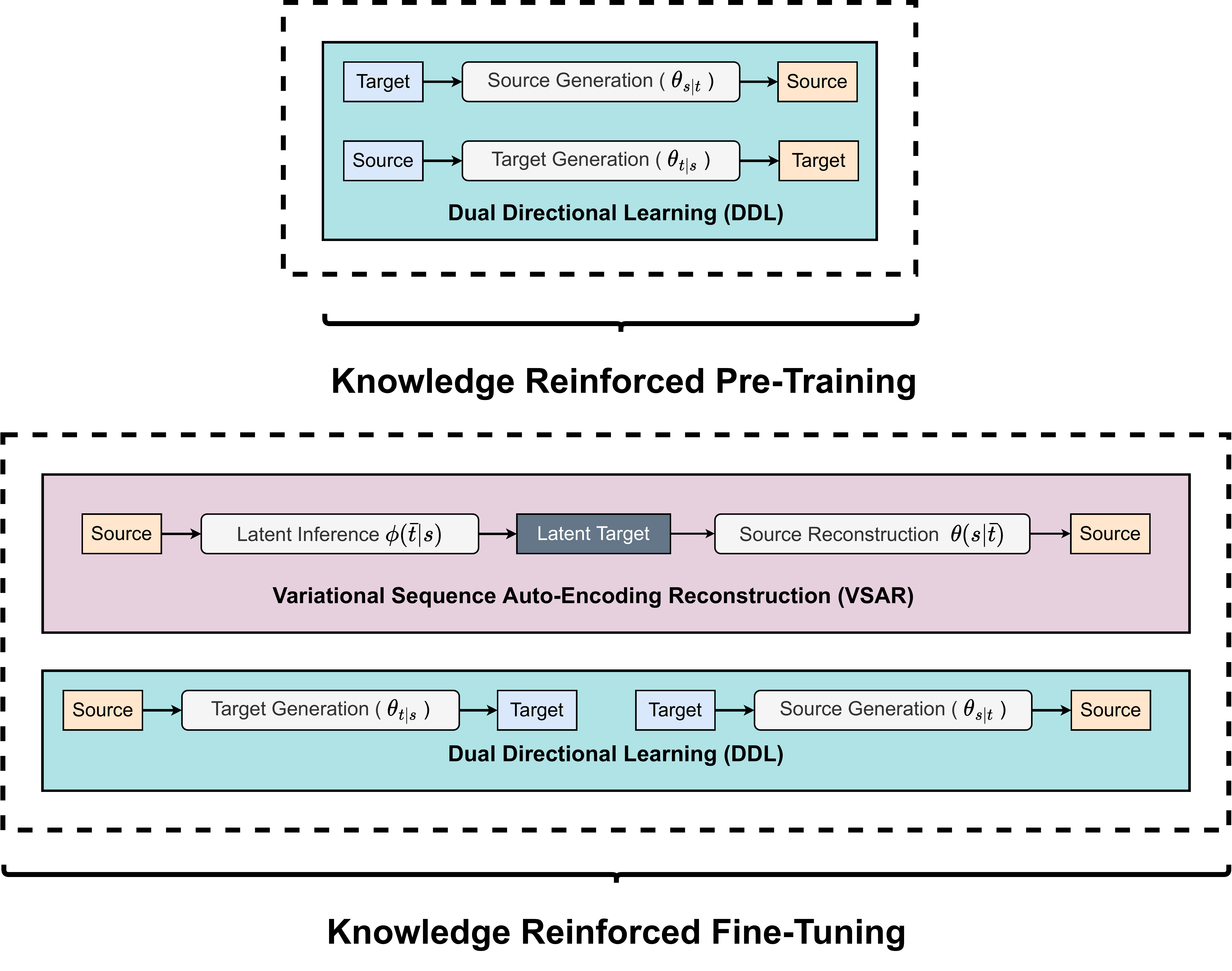}}
\caption{Knowledge Reinforced Learning (\textbf{KRL}).}
\label{fig: knowledge_reinforced_learning}
\end{figure*}

\section{Combining VSAR and DDL for Semi-supervised Learning}
\label{sec: vsar-and-ddl}

In this section, we introduce our semi-supervised learning model (VSAR+DDL), which combines models presented in previous sections. For semi-supervised learning, the log-likelihood of the data can be expressed as follow:

\begin{equation}
\begin{aligned}
    & \boldsymbol{L} = \boldsymbol{L_{1}} + \boldsymbol{L_{2}} \\ &= \sum_{i=1}^{N}\{\mathbb{E}_{q(\bar{t} \vert s_{(i)}; \boldsymbol{\phi})}[\log p(s_{(i)} \vert \bar{t}; \boldsymbol{\theta})] - \mathbb{D}_{KL}[q(\bar{t} \vert s_{(i)}; \boldsymbol{\phi}) \vert  \vert p(\bar{t})]\} \\ &+ \sum_{i=1}^{M}(\log p(s_{(i)} \vert t_{(i)}; \boldsymbol{\theta_{s \vert t}}) + \log p(t_{(i)} \vert s_{(i)}; \boldsymbol{\theta_{t \vert s}}))
\end{aligned}
\label{equ: vsar_ddl_likelihood}
\end{equation}

As suggested in equation \ref{equ: vsar_ddl_likelihood}, for unsupervised learning and supervised learning, the likelihood function involves the same set of conditional probability between $\boldsymbol{s}$ and $\boldsymbol{t}$. We hypothesise that sharing parameters between these two models is beneficial. {Thus,} we share two sets of neural network parameters from the VSAR and DDL models (i.e. $q_{\boldsymbol{\phi}}(\bar{\boldsymbol{t}} \vert \boldsymbol{s}) \equiv p_{\boldsymbol{\theta_{t \vert s}}}(\boldsymbol{t} \vert \boldsymbol{s})$ and $p_{\boldsymbol{\theta}}(\boldsymbol{s} \vert \bar{\boldsymbol{t}}) \equiv p_{\boldsymbol{\theta_{s \vert t}}}(\boldsymbol{s} \vert \boldsymbol{t})$). This allows the strong supervision signal from the DDL model to {contribute to the VSAR model, directly}. At the same time, the unsupervised signal from the VSAR model can benefit the generalisation of the DDL model.

\subsection{Knowledge Reinforced Learning}

Our empirical experiments suggest that our combined model (DDL+VSAR) suffers from a cold-start problem for parameter optimisation{,} when conducting semi-supervised learning from scratch. We found that a key to the success of our model is to have \textbf{better initialisation} of the model weight. Hence, we present a novel training scheme called \textbf{knowledge reinforced learning} (Figure \ref{fig: knowledge_reinforced_learning}), which includes two-stage training. In stage one (pre-training), we conduct supervised learning with our DDL model on paired training sets, as demonstrated in Algorithm \ref{alg: KRL-pretrain}. In stage two (fine-tuning), we initialise the VSAR model parameter with the best performance DDL model from stage one; and we conduct semi-supervised learning with labelled and unlabelled data, as demonstrated in Algorithm \ref{alg: KRL-finetuning}. The intuition is to inject better preliminary information into training the {VS}AR model.

\begin{algorithm}[!h]
\caption{Knowledge Reinforced Pre-Training}
\label{alg: KRL-pretrain}
\textbf{Input}: \\ Supervised Training Data ($\mathcal{D}^{S}_{T}=\{(s_{1}, t_{1}), ..., (s_{N}, t_{N})\}$), Supervised Validation Data ($\mathcal{D}^{S}_{V}$) \\
\textbf{Parameter}: \\ DDL Model: $\boldsymbol{\theta_{s \vert t}}$ and $\boldsymbol{\theta_{t \vert s}}$ \\
\textbf{Parameter Sharing}: \\ Set $\boldsymbol{\theta_{s \vert t}}$ equals to $\boldsymbol{\theta_{t \vert s}}$ {throughout} knowledge reinforced pre-training \\
\textbf{Output}: $\boldsymbol{\theta_{s \vert t}}^{*}$ and $\boldsymbol{\theta_{t \vert s}}^{*}$
\begin{algorithmic}[1] %[1] enables line numbers
\STATE Initialise $\boldsymbol{\theta_{s \vert t}}$ and $\boldsymbol{\theta_{t \vert s}}$ with a random seed; set maximum training epochs as $\boldsymbol{T}$; set $\boldsymbol{L_{2}}^{*}=0$
\WHILE{{m}aximum epochs not reached}
\STATE Update $\boldsymbol{\theta_{s \vert t}}$ and $\boldsymbol{\theta_{t \vert s}}$ with mini-batch data from $\mathcal{D}^{S}_{T}$ based on Equation \ref{equ: ddl_marginal_likelihood}
\IF {$\boldsymbol{L_{2}}$ in Equation \ref{equ: ddl_marginal_likelihood} calculated based on $\mathcal{D}^{S}_{V}$ {larger} than $\boldsymbol{L_{2}}^{*}$}
\STATE Set $\boldsymbol{L_{2}}^{*} \gets \boldsymbol{L_{2}}$
\STATE Set $\boldsymbol{\theta_{s \vert t}}^{*} \gets \boldsymbol{\theta_{s \vert t}}$
\STATE Set $\boldsymbol{\theta_{t \vert s}}^{*} \gets \boldsymbol{\theta_{t \vert s}}$
\ENDIF
\ENDWHILE \\
\textbf{Return}: $\boldsymbol{\theta_{s \vert t}}^{*}$ and $\boldsymbol{\theta_{t \vert s}}^{*}$
\end{algorithmic}
\end{algorithm}

\begin{algorithm}[!h]
\caption{Knowledge Reinforced Fine-Training}
\label{alg: KRL-finetuning}
\textbf{Input}: \\ Unsupervised Data ($\mathcal{D}^{U}=\{s_{1}, ..., s_{M}\}$) \\ Supervised Training Data ($\mathcal{D}^{S}_{T}=\{(s_{1}, t_{1}), ..., (s_{N}, t_{N})\}$), Supervised Validation Data ($\mathcal{D}^{S}_{V}$) \\
\textbf{Parameter}: \\ VSAR Model: $\boldsymbol{\phi}$ and $\boldsymbol{\theta}$; DDL Model: $\boldsymbol{\theta_{s \vert t}}$ and $\boldsymbol{\theta_{t \vert s}}$ \\
\textbf{Parameter Sharing}: \\ Set $\boldsymbol{\phi}$ equals to $\boldsymbol{\theta_{t \vert s}}$; $\boldsymbol{\theta}$ equals to $\boldsymbol{\theta_{s \vert t}}$; and $\boldsymbol{\theta_{s \vert t}}$ equals to $\boldsymbol{\theta_{t \vert s}}$ {throughout} knowledge reinforced fine-tuning \\
\textbf{Output}: $\boldsymbol{\theta_{s \vert t}}^{**}$, $\boldsymbol{\theta_{t \vert s}}^{**}$; $\boldsymbol{\phi}^{**}$ and $\boldsymbol{\theta}^{**}$
\begin{algorithmic}[1] %[1] enables line numbers
\STATE Initialise $\boldsymbol{\phi}$ and $\boldsymbol{\theta_{t \vert s}}$ with $\boldsymbol{\theta_{t \vert s}}^{*}$; and initialise $\boldsymbol{\theta}$ and $\boldsymbol{\theta_{s \vert t}}$ with $\boldsymbol{\theta_{s \vert t}}^{*}$; set maximum training epochs as $\boldsymbol{T}$; set $\boldsymbol{L_{2}}^{*}=0$.
\WHILE{{m}aximum epochs not reached}
\STATE Update $\boldsymbol{\theta_{s \vert t}}$ and $\boldsymbol{\theta_{t \vert s}}$ with mini-batch data from $\mathcal{D}^{S}_{T}$ based on Equation \ref{equ: ddl_marginal_likelihood}
\STATE Update $\boldsymbol{\phi}$ and $\boldsymbol{\theta}$ with mini-batch data from $\mathcal{D}^{U}$ based on Equation \ref{equ: vasr_marginal_likelihood}
\IF {$\boldsymbol{L_{2}}$ in Equation \ref{equ: ddl_marginal_likelihood} calculated based on $\mathcal{D}^{S}_{V}$ {larger} than $\boldsymbol{L_{2}}^{*}$}
\STATE Set $\boldsymbol{L_{2}}^{*} \gets \boldsymbol{L_{2}}$
\STATE Set $\boldsymbol{\theta_{s \vert t}}^{**} \gets \boldsymbol{\theta_{s \vert t}}$
\STATE Set $\boldsymbol{\theta_{t \vert s}}^{**} \gets \boldsymbol{\theta_{t \vert s}}$
\STATE Set $\boldsymbol{\phi}^{**} \gets \boldsymbol{\phi}$
\STATE Set $\boldsymbol{\theta}^{**} \gets \boldsymbol{\theta}$
\ENDIF
\ENDWHILE \\
\textbf{Return}: $\boldsymbol{\theta_{s \vert t}}^{**}$, $\boldsymbol{\theta_{t \vert s}}^{**}$; $\boldsymbol{\phi}^{**}$ and $\boldsymbol{\theta}^{**}$
\end{algorithmic}
\end{algorithm}

\subsection{Effect of Language Model Prior}

In literature \cite{higgins2016beta,miao2016language,yang2018unsupervised,he2020probabilistic}, a language model prior is introduced for regularisation purposes, which enforces samples to more likely contain a `reasonable' natural language, especially at the beginning of the training. Hence, we adopt the same approach and use a prior in our model. We empirically found the prior useful when the labelled dataset {wa}s relatively small. However, surprisingly, we found that training without a prior in the VSAR model yields better results {with our parameter initialisation method,} when the dataset is large. The improvement is significant ($p <.05$; Wilcoxon test), as shown in Table \ref{tab: main_quora} and Table \ref{tab: main_mscoco}. We report the results without language model prior as DDL +VSAR$^{\ast}$, and the log-likelihood becomes:

\begin{equation}
\begin{aligned}
    & \boldsymbol{L}^{\ast} = \sum_{i=1}^{N}\{\mathbb{E}_{q(\bar{t} \vert s_{(i)}; \boldsymbol{\phi})}[\log p(s_{(i)} \vert \bar{t}; \boldsymbol{\theta})] \} \\ &+ \sum_{i=1}^{M}(\log p(s_{(i)} \vert t_{(i)}; \boldsymbol{\theta_{s \vert t}}) + \log p(t_{(i)} \vert s_{(i)}; \boldsymbol{\theta_{t \vert s}}))
\end{aligned}
\label{equ: vsar_ddl_likelihood_no_prior}
\end{equation}

To further investigate this issue, we conducted experiments to compare the performance of semi-supervised learning when training with {prior} {(Eq.} \ref{equ: vsar_ddl_likelihood}{)} {or without prior ( Eq.} \ref{equ: vsar_ddl_likelihood_no_prior}{)} under different data portion setting. We empirically found that with a low portion of labelled data, the combined model (DDL+VSAR) with a prior grant{s} significantly ($p <.05$; Wilcoxon test) better performance and is more stable. This aligns with the observations in \cite{higgins2016beta,miao2016language,yang2018unsupervised,he2020probabilistic}. However, with a large portion of labelled data, the combined model (DDL+VSAR) without the prior is significantly ($p <.05$; Wilcoxon test) better. 

We argue that this phenomenon relates to our choice of  prior{,} as it is pre-trained on {an} unlabelled source text corpus{,} instead of on the target text corpus. This approximation leads to a distribution shift from the true prior distribution $p(\bar{\boldsymbol{t}})$. Thus, when a low portion of {the} labelled data is used in Algorithm \ref{alg: KRL-pretrain}, the final DDL parameters $\boldsymbol{\theta_{s \vert t}}^{*}$ and $\boldsymbol{\theta_{t \vert s}}^{*}$ for {the} initialisation VSAR model in  Algorithm \ref{alg: KRL-finetuning} {are} not good enough. The prior{, in this case,} can still benefit the combined model in the semi-supervised learning setting. However, with a large portion of labelled data, the initialisation is good enough, and{, in such a case,} the distribution shift can harm the combined model.

\subsection{Semi-supervised Learning Setup}

Under the semi-supervised learning setting, we limit the size of the supervised source and target pairs to be less than or equal to the unsupervised source text ($M \leq N$), as we could otherwise just conduct supervised learning{,} to take full advantage of {the} observed data pairs. This paper presents a thorough study {of} different sizes {for} $M$ and $N$. Experimental results under this setting are presented in Table \ref{tab: semi_quora}, Table \ref{tab: semi_mscoco}, Table \ref{tab: semi_parabank} and Table \ref{tab: semi_paramnt}.

\section{Experiments}

Here, we describe the datasets, experimental setup, evaluation metrics and experimental results.

\subsection{Datasets}

\textbf{MSCOCO} \cite{lin2014microsoft}: This dataset has been widely adopted to evaluate paraphrase generation methods and contains human-annotated captions of images. Each image is associated with five captions from different annotators, who describe the most prominent object or action in an image. We use the $2017$ version for our experiments; from the five captions accompanying each image, we randomly choose one as the source string and one as the target string for training. We randomly choose one as the source string for testing and use the rest four as the references.

\textbf{Quora}\footnote{https://quoradata.quora.com/First-Quora-Dataset-Release-Question-Pairs}: This dataset consists of $150$K lines of question duplicate pairs, and it has been used as a benchmark dataset for paraphrase generation since $2017$. However, since this dataset does not contain a specific split for training and testing, prior models are evaluated based on different subset sizes of data.

For both datasets (MSCOCO and Quora), in order to improve {the reproducibility} of our results, we use a pre-trained tokeni{s}er ('bert-base-uncased' version) from \cite{devlin2018bert}\footnote{https://github.com/huggingface/transformers} and set the maximum token length as $20$ (by removing the tokens beyond the first $20$). Following \cite{li2019decomposable,fu2020paraphrase,su2021keep}, we use training, validation and test sets as $100$K, $4$K and $20$K{, respectively} for {the} Quora dataset; and $93$K, $4$K and $20$K{, respectively,} for MSCOCO. For the complementary study in Table \ref{tab: complement_quora} and Table \ref{tab: complement_mscoco}, we use training, validation and test sets as $100$K, $24$K and $24$K for {the} Quora dataset; and $100$K, $5$K and $5$K for MSCOCO, in order to have a fair comparison with the results reported in \cite{hosking2021factorising,hosking2022hierarchical}.

{Other available datasets for paraphrase generation tasks include: ParaBank} \cite{hu2019parabank} {and PARANMT} \cite{wieting2017paranmt} {, which are two large-scale datasets created using back translation techniques from another non-English parallel corpus. Since these two datasets are less adopted by researchers in the literature, we can not directly compare them against existing works. Still, as an alternative, to further demonstrate the efficacy of our proposed models, we conduct semi-supervised learning experiments on these two datasets. For the ParaBank dataset, we took the 'ParaBank v1.0 (5m pairs)'\footnote{https://nlp.jhu.edu/parabank/} and use a similar setup as for the Quora dataset. The dataset consists of 5M lines of duplicated pairs; we randomly choose the same subset of the data for our experiments and use training, validation and test set of size 100K, 4K and 20K, respectively. For the PARANMT dataset, we first filter out the paraphrase pairs (remove entries where the paragram-phrase score\footnote{The paragram-phrase score measures the semantic similarity between a pair of sentences. For the complete PARANMT dataset, the mean score is 0.69, with a standard deviation of 0.26. We select the range among one standard deviation of the mean, i.e. (0.43, 0.95). Since we wish the dataset to be in an actual paraphrase form, we further limit the lower bound to 0.9, to ensure the quality of the data.} is higher than 0.95 and smaller than 0.90; set the maximum token length as 20 and the minimum token length as 5) and keep the middle percentiles, as recommended in} \cite{wieting2017paranmt}{, to remove noisy and trivial paraphrases. After the filtering, we apply a similar experimental setup as the Quora dataset and randomly choose the same subset of the data for our experiments and use training, validation and test set of size 100K, 4K and 20K, respectively.}  

\subsection{Baselines}

We consider several state-of-the-art baselines, presented in Table \ref{tab: main_quora}, Table \ref{tab: main_mscoco}, Table \ref{tab: complement_quora}, and Table  \ref{tab: complement_mscoco}. Note that these experimental results are directly taken from \cite{su2021keep}\footnote{The authors do not make their code publicly available.} and \cite{hosking2022hierarchical}. For evaluation, we start with our implementation of the Transformer model as the absolute baseline, which achieves competitive performance{,} as reported in \cite{su2021keep}. The Transformer model \cite{vaswani2017attention} is {considered} as the SOTA model, which is very `hard to beat'. We report our model performance based on a similar setup as in \cite{su2021keep} and \cite{hosking2022hierarchical}.

{Recently, large-scale pre-trained language models (PLMs) have been widely adopted as the state-of-the-art approaches for both understanding and generation tasks in the NLP domain; in this paper, however, PLMs are not selected as the baseline model to compete against, as they contain external information trained in an unsupervised fashion based on a large-scale text corpus. Alternatively, in this paper, we focused on end-to-end learning for paraphrase generation tasks from scratch. This experimental setting allows us to directly compare with other literature. Hence, in this paper, we are not compared against any PLMs, other than leaving this as future work. However, note that all types of PLM models are primarily based on the Transformer architecture, which we used as a baseline model to compare with. The implication is that our methods could potentially be used for improving the performance of PLMs with limited labelled resources and large-scale unlabelled data points.}

\begin{table*}[!ht]
% \smallskip
% \centering
\caption{\bf Semi-Supervised Learning Experiment Results for Quora.}
\resizebox{0.9\columnwidth}{!}{
% \begin{adjustwidth}{-2.25in}{0in}{
\centering
\smallskip
\begin{tabular}{lcccccccccc}
\hline
Model & Labelled & Unlabelled & B-$1$ & B-$2$ & B-$3$ & B-$4$ & i-B & R-$1$ & R-$2$ & R-L\\
\hline
% Transformer & $20$K & $-$ & $69.39$ & $50.17$ & $36.06$ & $26.49$ & $18.43$ & $42.08$ & $16.31$ & $38.27$ \\
DDL & $20$K & $-$ & $46.68$ & $33.44$ & $25.46$ & $20.18$ & $11.08$ & $47.57$ & $25.42$ & $45.50$ \\
DDL+VSAR$_{1}$ & $20$K & $20$K & $47.80\uparrow$ & $34.33\uparrow$ & $26.17\uparrow$ & $20.76\uparrow$ & $11.25\uparrow$ & $48.03\uparrow$ & $25.82\uparrow$ & $45.84\uparrow$ \\
DDL+VSAR$_{2}$ & $20$K & $100$K & $\textbf{50.26}\uparrow$ & $\textbf{36.87}\uparrow$ & $\textbf{28.50}\uparrow$ & $\textbf{22.82}\uparrow$ & $\textbf{11.60}\uparrow$ & $\textbf{51.51}\uparrow$ & $\textbf{28.45}\uparrow$ & $\textbf{49.07}\uparrow$ \\
\hline
% Transformer & $50$K & $-$ & $69.39$ & $50.17$ & $36.06$ & $26.49$ & $18.43$ & $42.08$ & $16.31$ & $38.27$ \\
DDL & $50$K & $-$ & $53.31$ & $40.22$ & $31.70$ & $25.80$ & $\textbf{13.80}$ & $\textbf{55.63}$ & $\textbf{32.15}$ & $\textbf{53.13}$ \\
DDL+VSAR$_{1}$ & $50$K & $50$K & $53.33\uparrow$ & $39.93\downarrow$ & $31.39\downarrow$ & $25.49\downarrow$ & $13.45\downarrow$ & $55.51\downarrow$ & $31.90\downarrow$ & $52.95\downarrow$ \\
DDL+VSAR$_{2}$ & $50$K & $100$K & $\textbf{53.79}\uparrow$ & $\textbf{40.47}\uparrow$ & $\textbf{31.86}\uparrow$ & $\textbf{25.93}\uparrow$ & $13.67\downarrow$ & $55.58\downarrow$ & $31.89\downarrow$ & $52.93\downarrow$ \\
% \hline
% DDL & $100$K & $-$ & 55.92 & 42.97 & 34.27 & 28.13 & 14.77 & 56.65 & 33.36 & 54.31 \\
% DDL+VSAR_{1} & $100$K & $100$K & B-1 & B-2 & B-3 & B-4 & i-B & R-1 & R-2 & R-L \\
\hline
\end{tabular}}
\label{tab: semi_quora}
% \end{adjustwidth}
\end{table*}

\begin{table*}[!ht]
% \smallskip
% \centering
\caption{\bf Semi-Supervised Learning Experiment Results for MSCOCO.}
\resizebox{0.9\columnwidth}{!}{
% \begin{adjustwidth}{-2.25in}{0in}{
\centering
\smallskip\begin{tabular}{lcccccccccc}
\hline
Model & Labelled & Unlabelled & B-1 & B-2 & B-3 & B-4 & i-B & R-1 & R-2 & R-L\\
\hline
% \hline
% Transformer & $20$K & $-$ & $69.39$ & $50.17$ & $36.06$ & $26.49$ & $18.43$ & $42.08$ & $16.31$ & $38.27$ \\
DDL & $20$K & $-$ & $66.82$ & $47.25$ & $33.14$ & $23.75$ & $16.66$ & $40.53$ & $14.95$ & $36.94$ \\
DDL+VSAR$_{1}$ & $20$K & $20$K & $66.98\uparrow$ & $47.28\uparrow$ & $33.10\downarrow$ & $23.72\downarrow$ & $16.54\downarrow$ & $40.60\uparrow$ & $14.95\uparrow$ & $36.94\uparrow$ \\
DDL+VSAR$_{2}$ & $20$K & $93$K & $\textbf{67.64}\uparrow$ & $\textbf{48.00}\uparrow$ & $\textbf{33.96}\uparrow$ & $\textbf{24.55}\uparrow$ & $\textbf{16.68}\uparrow$ & $\textbf{40.87}\uparrow$ & $\textbf{15.12}\uparrow$ & $\textbf{37.01}\uparrow$ \\
\hline
% Transformer & $50$K & $-$ & $69.39$ & $50.17$ & $36.06$ & $26.49$ & $18.43$ & $42.08$ & $16.31$ & $38.27$ \\
DDL & $50$K & $-$ & $69.39$ & $50.17$ & $36.06$ & $26.49$ & $18.43$ & $42.08$ & $16.31$ & $38.27$ \\
DDL+VSAR$_{1}$ & $50$K & $50$K & $69.43\uparrow$ & $50.21\uparrow$ & $36.08\uparrow$ & $26.45\downarrow$ & $18.31\downarrow$ & $42.20\uparrow$ & $16.33\uparrow$ & $38.31\uparrow$ \\
DDL+VSAR$_{2}$ & $50$K & $93$K & $\textbf{69.91}\uparrow$ & $\textbf{50.65}\uparrow$ & $\textbf{36.52}\uparrow$ & $\textbf{26.93}\uparrow$ & $\textbf{18.51}\uparrow$ & $\textbf{42.39}\uparrow$ & $\textbf{16.46}\uparrow$ & $\textbf{38.40}\uparrow$ \\
% DDL & $93$K & $-$ & $70.81$ & $51.76$ & $37.69$ & $28.04$ & $19.43$ & $41.67$ & $16.55$ & $38.14$ \\
% DDL+VSAR & $93$K & $93$K & B-1 & B-2 & B-3 & B-4 & i-B & R-1 & R-2 & R-L \\
\hline
\end{tabular}}
\label{tab: semi_mscoco}
% \end{adjustwidth}
\end{table*}

\begin{table*}[!ht]
% \smallskip
% \centering
\caption{\bf Semi-Supervised Learning Experiment Results for ParaBank.}
\resizebox{0.9\columnwidth}{!}{
% \begin{adjustwidth}{-2.25in}{0in}{
\centering
\smallskip
\begin{tabular}{lcccccccccc}
\hline
Model & Labelled & Unlabelled & B-$1$ & B-$2$ & B-$3$ & B-$4$ & i-B & R-$1$ & R-$2$ & R-L\\
\hline
% Transformer & $20$K & $-$ & $69.39$ & $50.17$ & $36.06$ & $26.49$ & $18.43$ & $42.08$ & $16.31$ & $38.27$ \\
DDL & $20$K & $-$ & $50.48$ & $39.21$ & $31.11$ & $25.16$ & $12.98$ & $57.03$ & $33.77$ & $55.31$ \\
DDL+VSAR$_{1}$ & $20$K & $20$K & $50.69\uparrow$ & $39.41\uparrow$ & $31.35\uparrow$ & $25.45\uparrow$ & $13.14\uparrow$ & $56.99\downarrow$ & $33.92\uparrow$ & $55.20\downarrow$ \\
DDL+VSAR$_{2}$ & $20$K & $100$K & $\textbf{54.04}\uparrow$ & $\textbf{43.41}\uparrow$ & $\textbf{35.44}\uparrow$ & $\textbf{29.40}\uparrow$ & $\textbf{14.94}\uparrow$ & $\textbf{60.66}\uparrow$ & $\textbf{37.50}\uparrow$ & $\textbf{58.90}\uparrow$ \\
\hline
% Transformer & $50$K & $-$ & $69.39$ & $50.17$ & $36.06$ & $26.49$ & $18.43$ & $42.08$ & $16.31$ & $38.27$ \\
DDL & $50$K & $-$ & $58.05$ & $48.84$ & $41.59$ & $35.82$ & $18.19$ & $66.45$ & $44.63$ & $64.81$ \\
DDL+VSAR$_{1}$ & $50$K & $50$K & $58.29\uparrow$ & $49.24\uparrow$ & $42.10\uparrow$ & $36.40\uparrow$ & $18.45\uparrow$ & $66.88\uparrow$ & $45.23\uparrow$ & $65.23\uparrow$ \\
DDL+VSAR$_{2}$ & $50$K & $100$K & $\textbf{59.22}\uparrow$ & $\textbf{50.33}\uparrow$ & $\textbf{43.23}\uparrow$ & $\textbf{37.51}\uparrow$ & $\textbf{18.89}\uparrow$ & $\textbf{67.64}\uparrow$ & $\textbf{46.05}\uparrow$ & $\textbf{65.98}\uparrow$ \\
% \hline
% DDL & $100$K & $-$ & 55.92 & 42.97 & 34.27 & 28.13 & 14.77 & 56.65 & 33.36 & 54.31 \\
% DDL+VSAR_{1} & $100$K & $100$K & B-1 & B-2 & B-3 & B-4 & i-B & R-1 & R-2 & R-L \\
\hline
\end{tabular}}
\label{tab: semi_parabank}
% \end{adjustwidth}
\end{table*}

\begin{table*}[!ht]
% \smallskip
% \centering
\caption{\bf Semi-Supervised Learning Experiment Results for PARANMT.}
\resizebox{0.9\columnwidth}{!}{
% \begin{adjustwidth}{-2.25in}{0in}{
\centering
\smallskip
\begin{tabular}{lcccccccccc}
\hline
Model & Labelled & Unlabelled & B-$1$ & B-$2$ & B-$3$ & B-$4$ & i-B & R-$1$ & R-$2$ & R-L\\
\hline
% Transformer & $20$K & $-$ & $69.39$ & $50.17$ & $36.06$ & $26.49$ & $18.43$ & $42.08$ & $16.31$ & $38.27$ \\
DDL & $20$K & $-$ & $62.49$ & $49.47$ & $39.92$ & $32.84$ & $17.40$ & $63.13$ & $40.43$ & $61.15$ \\
DDL+VSAR$_{1}$ & $20$K & $20$K & $63.48\uparrow$ & $50.70\uparrow$ & $41.32\uparrow$ & $34.33\uparrow$ & $18.16\uparrow$ & $64.59\uparrow$ & $41.97\uparrow$ & $62.58\uparrow$ \\
DDL+VSAR$_{2}$ & $20$K & $100$K & $\textbf{66.46}\uparrow$ & $\textbf{54.20}\uparrow$ & $\textbf{44.80}\uparrow$ & $\textbf{37.60}\uparrow$ & $\textbf{19.42}\uparrow$ & $\textbf{67.55}\uparrow$ & $\textbf{45.25}\uparrow$ & $\textbf{65.49}\uparrow$ \\
\hline
% Transformer & $50$K & $-$ & $69.39$ & $50.17$ & $36.06$ & $26.49$ & $18.43$ & $42.08$ & $16.31$ & $38.27$ \\
DDL & $50$K & $-$ & $70.55$ & $59.53$ & $50.79$ & $43.78$ & ${22.82}$ & ${71.97}$ & ${51.30}$ & ${69.97}$ \\
DDL+VSAR$_{1}$ & $50$K & $50$K & $70.41\downarrow$ & $59.48\downarrow$ & $50.81\uparrow$ & $43.87\uparrow$ & $23.02\uparrow$ & $72.06\uparrow$ & $51.45\uparrow$ & $70.07\uparrow$ \\
DDL+VSAR$_{2}$ & $50$K & $100$K & $\textbf{70.91}\uparrow$ & $\textbf{60.12}\uparrow$ & $\textbf{51.48}\uparrow$ & $\textbf{44.53}\uparrow$ & $\textbf{23.25}\uparrow$ & $\textbf{72.57}\uparrow$ & $\textbf{52.03}\uparrow$ & $\textbf{70.57}\uparrow$ \\
% \hline
% DDL & $100$K & $-$ & 55.92 & 42.97 & 34.27 & 28.13 & 14.77 & 56.65 & 33.36 & 54.31 \\
% DDL+VSAR_{1} & $100$K & $100$K & B-1 & B-2 & B-3 & B-4 & i-B & R-1 & R-2 & R-L \\
\hline
\end{tabular}}
\label{tab: semi_paramnt}
% \end{adjustwidth}
\end{table*}

\subsection{Experimental Setup}

In this section, we introduce our primary experimental setup. We do not use any external word embedding{,} such as Glove \cite{pennington2014glove}, word2vec\cite{mikolov2013distributed} or BERT \cite{devlin2018bert} for initialisation; rather, we obtain word embedding with end-to-end training, in order not to use any prior knowledge and better understand the impact of our model. We use the `base' version of the Transformer model \cite{vaswani2017attention}, which is a $6$-layer model with $512$ hidden units and $8$ heads for each encoder and decoder network. In each encoder and decoder, we have a separate learnable position embedding and its associated word embedding component.

We use a greedy decoding scheme for paraphrase generation, which is fast and cheap to compute. For model optimisation, we use Adam \cite{kingma2014adam} as our optimiser with default hyper-parameters ($\beta_{1}=0.9$, $\beta_{2}=0.999$, $\epsilon=1e-8$). We conduct all the experiments with a batch size of $512$ for the Quora and MSCOCO datasets. We set the learning rate as $1e-4$ for MSCOCO and $2e-4$ for Quora based on empirical experiments. All experiments are run for a maximum of $30$ epochs on NVidia GPU Cluster with A100 GPU. Experiments are repeated three times{,} with different random seeds ($1000$, $2000$ and $3000$){,} and the average result is reported in Tables $1$-$6$.

\begin{table*}[!h]
% \smallskip
% \centering
\caption{\textbf{Experiment Results for Quora.}}
\resizebox{0.9\columnwidth}{!}{
% \begin{adjustwidth}{-2.25in}{0in}{
\centering
% {
\smallskip\begin{tabular}{lcccccccc}
\hline
Model & B-1 & B-2 & B-3 & B-4 & i-B & R-1 & R-2 & R-L\\
\hline
Upper Bound (Copy Source) & $63.36$ & $49.99$ & $40.47$ & $33.54$ & - & $63.04$ & $38.15$ & $59.64$ \\
Lower Bound (Random Select)& $16.10$ & $4.50$ & $1.94$ & $0.79$ & - & $9.13$ & $1.54$ & $8.79$ \\
\hline
Residual-LSTM \cite{prakash2016neural} & $53.59$ & $39.49$ & $30.25$ & $23.69$ & $15.93$ & $55.10$ & $33.86$ & $53.61$ \\
$\beta$-VAE \cite{higgins2016beta} & $47.86$ & $33.21$ & $24.96$ & $19.73$ & $10.28$ & $47.62$ & $25.49$ & $45.46$ \\
Transformer \cite{vaswani2017attention} & $53.56$ & $40.47$ & $32.11$ & $25.01$ & $17.98$ & $57.82$ & $32.58$ & $56.26$ \\
LBOW-TOPk \cite{fu2020paraphrase} & $55.79$ & $42.03$ & $32.71$ & $26.17$ & $19.03$ & $58.79$ & $34.57$ & $56.43$ \\
IANet+X \cite{su2021keep} & $56.06$ & $42.69$ & $33.38$ & $26.52$ & $19.62$ & $59.33$ & $35.01$ & $57.13$ \\
\hline
Transformer (our implementation) & $54.73$ & $41.59$ & $32.96$ & $26.94$ & $14.50$ & $56.90$ & $33.28$ & $54.29$ \\
DDL (our model) & $55.97\uparrow$ & $43.02\uparrow$ & $34.32\uparrow$ & $28.19\uparrow$ & $14.83\uparrow$ & $\textbf{58.80}\uparrow$ & $35.00\uparrow$ & $56.11\uparrow$ \\
DDL + SVAR (our model) & $55.79\uparrow$ & $42.79\uparrow$ & $34.11\uparrow$ & $28.01\uparrow$ & $\textbf{14.92}\uparrow$ & $58.61\uparrow$ & $34.75\uparrow$ & $55.91\uparrow$ \\
DDL + SVAR$^{\ast}$ (our model) & $\textbf{55.99}\uparrow$ & $\textbf{43.05}\uparrow$ & $\textbf{34.37}\uparrow$ & $\textbf{28.23}\uparrow$ & $14.81\uparrow$ & $58.79\uparrow$ & $\textbf{35.02}\uparrow$ & $\textbf{56.14}\uparrow$ \\
\hline
\end{tabular}}
\label{tab: main_quora}
% \end{adjustwidth}
\end{table*}

\begin{table*}[!h]
% \smallskip
% \centering
\caption{\textbf{Experiment Results for MSCOCO.}}
\resizebox{0.9\columnwidth}{!}{
% \begin{adjustwidth}{-2.25in}{0in}{
\centering
% {
\smallskip\begin{tabular}{lcccccccc}
\hline
Model & B-1 & B-2 & B-3 & B-4 & i-B & R-1 & R-2 & R-L\\
\hline
Upper Bound (Copy Source) & $64.97$ & $44.90$ & $30.69$ & $21.30$ & - & $39.18$ & $12.96$ & $34.61$ \\
Lower Bound (Random Select)& $32.34$ & $10.99$ & $3.81$ & $1.68$ & - & $17.58$ & $1.51$ & $16.27$ \\
\hline
Residual-LSTM \cite{prakash2016neural} & $70.24$ & $48.65$ & $34.04$ & $23.66$ & $18.72$ & $41.07$ & $15.26$ & $37.35$ \\
$\beta$-VAE \cite{higgins2016beta} & $70.04$ & $47.59$ & $32.29$ & $22.54$ & $18.34$ & $40.72$ & $14.75$ & $36.75$ \\
Transformer \cite{vaswani2017attention} & $71.31$ & $49.86$ & $35.55$ & $24.68$ & $19.81$ & $41.49$ & $15.84$ & $37.09$ \\
LBOW-TOPk \cite{fu2020paraphrase} & $72.60$ & $51.14$ & $35.66$ & $25.27$ & $21.07$ & $42.08$ & $16.13$ & $38.16$ \\
IANet+X \cite{su2021keep} & $72.10$ & $52.22$ & $37.39$ & $26.06$ & $21.28$ & $43.81$ & $16.35$ & $39.65$ \\
\hline
Transformer (our implementation) & $68.72$ & $49.64$ & $35.87$ & $26.63$ & $18.59$ & $42.09$ & $16.53$ & $38.35$ \\
DDL (our model) & $70.75\uparrow$ & $51.72\uparrow$ & $37.62\uparrow$ & $27.95\uparrow$ & $19.37\uparrow$ & $43.00\uparrow$ & $17.01\uparrow$ & $39.06\uparrow$ \\
DDL + SVAR (our model) & $70.84\uparrow$ & $51.84\uparrow$ & $37.75\uparrow$ & $28.04\uparrow$ & $19.39\uparrow$ & $\textbf{43.05}\uparrow$ & $\textbf{17.04}\uparrow$ & $\textbf{39.07}\uparrow$ \\
DDL + SVAR$^{\ast}$ (our model) & $\textbf{70.99}\uparrow$ & $\textbf{51.91}\uparrow$ & $\textbf{37.82}\uparrow$ & $\textbf{28.12}\uparrow$ & $\textbf{19.39}\uparrow$ & $43.00\uparrow$ & $17.03\uparrow$ & $39.02\uparrow$ \\
\hline
\end{tabular}}
\label{tab: main_mscoco}
% \end{adjustwidth}
\end{table*}

\subsection{Evaluation}

In this paper, we evaluate our models {first} based on \textit{quantitative metrics}: BLEU \cite{papineni2002bleu}\footnote{https://www.nltk.org/}, ROUGE \cite{lin2004rouge}\footnote{https://github.com/huggingface/datasets/tree/master/metrics/rouge}, and i-BLEU \cite{sun2012joint}. {The main justification behind this is to compare with existing work in the literature directly, which focused on end-to-end learning of paraphrase generation tasks from scratch, using deep neural network models.} BLEU (Bilingual Evaluation Understudy) and ROUGE (Recall-Oriented Understudy for Gisting Evaluation) scores are based on `n-gram' coverage between system-generated paraphrase(s) and reference sentences. They have been used widely{,} to automatically evaluate the quality and accuracy of natural language generation tasks.

Previous work has shown that automatic evaluation metrics can perform well for paraphrase identification tasks \cite{madnani2012re} and correlate well with human judgements in evaluating generated paraphrases \cite{wubben2010paraphrase}. Recent papers introduce additional i-BLEU \cite{sun2012joint} metrics{,} to balance the fidelity of generated outputs to reference paraphrases (BLEU){,} as well as the level of diversity introduced (self-B). For all metrics apart from self-B, the higher the value, the better the model performs.

{Additionally, we present \textit{qualitative evaluation} results in Tables} \ref{tab:supervised_qualitative} and \ref{tab:semisupervised_qualitative}, {based on the Quora dataset. Our qualitative evaluation aims to examine two aspects, as follows: (1) how does our proposed supervised model (DDL)  compare with the very strong supervised learning baseline Transformer, given different sizes of labelled pairs data sets; and (2) what are the benefits of our proposed semi-supervised model (DDL+VSAR) when incorporating more unlabelled data, given the labelled pairs data set size remains constant.}  

% in The main justification behind this is to compare with existing work in the literature directly, which focused on end-to-end learning of paraphrase generation task from a machine learning perspective.

\subsection{Results and Discussion}
\label{sec: results_and_discussion}

\subsubsection{Learning with Unlabelled Data Only}
\label{sec: results_unlabelled_data_only}

{In an initial experiment, we explored the ability of the VSAR model to perform paraphrase generation tasks using only unlabelled data. This experiment was conducted to see if the model could accurately capture the information required for paraphrase generation, without the aid of labelled data. However, the results of our initial experiment showed that the VSAR model alone was not able to produce high-quality paraphrases, and often resulted in sentences that were either incomprehensible or meaningless. These results motivated us to pursue a semi-supervised learning solution for paraphrase generation, which would provide the model with guidance from labelled data. For unsupervised learning with VSAR, we found that while the lower bound indicated by $\boldsymbol{L}_{1}$ (Equation} \ref{equ: vasr_marginal_likelihood}) {decreased during training in a fully unsupervised setting, the model still generated low-quality paraphrases. This further validated the need for a semi-supervised learning solution, which is introduced in section} \ref{sec: vsar-and-ddl} {of this paper.}

\subsubsection{Learning with a Fraction of Data}

In this section, we present results which are based on a fraction of labelled data in Table{s} \ref{tab: semi_quora}, \ref{tab: semi_mscoco} \ref{tab: semi_parabank} and \ref{tab: semi_paramnt}. In {all four} tables, we present the results of two models - the supervised learning model, DDL and the semi-supervised learning model, DDL + VSAR. In a semi-supervised learning setting, VSAR is trained on unlabelled data, and DDL is trained on labelled data. The DDL+VSAR$_{1}$ model employs equivalent sized labelled and unlabelled datasets, which come from the same source and target pairs, so there is no additional information applied in this case. The DDL+VSAR$_{2}$ model employs the full unlabelled dataset in addition to the existing labelled dataset, which is the true semi-supervised setting.

Results suggest that the DDL+VSAR$_{1}$ model achieves competitive or better performance on most metrics' scores compared to the supervised DDL model only trained on labelled data; especially with a lower fraction of the data (for example, the significant improvement for $20$K is more noticeable than for $50$K). Furthermore, fixing the labelled data size, the DDL+VSAR$_{2}$ model achieves significantly better performance by using additional unlabelled data{,} than all other models reported in both tables ($p <.05$; Wilcoxon test), which means the semi-supervised learning does work in this scenario.

\begin{table}[!h]
% \smallskip
% \centering
\caption{\textbf{Complement Results for Quora.}}
\resizebox{0.70\columnwidth}{!}{
% \begin{adjustwidth}{-0.70in}{0in}{
\centering
% {
\smallskip\begin{tabular}{lccc}
\hline
Model & B-4 & self-B & i-B\\
% \hline
% Upper Bound (Copy Source) & 63.36 & 49.99 & 40.47  \\
% Lower Bound (Random Select)& 16.10 & 4.50 & 1.94 \\
\hline
Separator \cite{hosking2021factorising}  & 23.68 & 24.20 & 14.10 \\
HRQ-VAE \cite{hosking2022hierarchical}  & 33.11 & 40.35 & 18.42 \\
\hline
Transformer (our implementation) & 26.92 & \textbf{35.33} & 14.47 \\
% DDL (our model) & $28.23\uparrow$ & $38.54\downarrow$ & $\textbf{14.87}\uparrow$ \\
DDL + SVAR (our model) & $28.15\uparrow$ & $38.92\downarrow$ & $\textbf{14.73}\uparrow$ \\
DDL + SVAR$^{\ast}$ (our model) & $\textbf{28.16}\uparrow$ & $39.07\downarrow$ & $14.71\uparrow$ \\
\hline
\end{tabular}}
\label{tab: complement_quora}
% \end{adjustwidth}
\end{table}

\begin{table}[!h]
% \smallskip
% \centering
\caption{\bf Complement Results for MSCOCO.}
\resizebox{0.70\columnwidth}{!}{
% \begin{adjustwidth}{-0.70in}{0in}{
\centering
% {
\smallskip\begin{tabular}{lccc}
\hline
Model & B-4 & self-B & i-B \\
% \hline
% Upper Bound (Copy Source) & 63.36 & 49.99 & 40.47  \\
% Lower Bound (Random Select)& 16.10 & 4.50 & 1.94 \\
\hline
Separator \cite{hosking2021factorising}  & $20.59$ & $12.76$ & $13.92$ \\
HRQ-VAE \cite{hosking2022hierarchical}  & $27.90$ & $16.58$ & $19.04$ \\
\hline
Transformer (our implementation) & $26.87$ & $\textbf{13.50}$ & $18.79$ \\
% DDL (our model) & $27.88\uparrow$ & $14.83\downarrow$ & $\textbf{19.34}\uparrow$ \\
DDL + SVAR (our model) & $27.87\uparrow$ & $15.42\downarrow$ & $19.21\uparrow$ \\
DDL + SVAR$^{\ast}$ (our model) & $\textbf{27.92}\uparrow$ & $15.21\downarrow$ & $\textbf{19.29}\uparrow$ \\
\hline
\end{tabular}}
\label{tab: complement_mscoco}
% \end{adjustwidth}
\end{table}

\begin{table}[h!]
\smallskip
\centering
\caption{{\textbf{Selected paraphrase generation results for Transformer (TRANS) versus DDL model with different amounts of labelled data (denoted in brackets), represented in the case of Quora dataset.}}}
\label{tab:supervised_qualitative}
\resizebox{.99\linewidth}{!}{
% \begin{adjustwidth}{-2.25in}{0in}{
\smallskip\begin{tabular}{lp{0.80\linewidth}}
\hline
\hline
Source: & what are best courses for journalism ? \\
\hline
Reference: & what are the best courses on journalism ? \\
\hline
\hline
TRANS (20K) & which is the best software for beginner ?\\
\hline
DDL (20K) & what is the best digital marketing course ?\\
\hline
TRANS (50K) & what are the best courses on nagpur ?\\
\hline
DDL (50K) & what are the best courses for journalism ?\\
\hline
TRANS (100K) & what are the best courses about journalism ?\\
\hline
DDL (100K) & what are the best courses for journalism ?\\
\hline
\hline
Source: & what helps asthma without an inhaler ?\\
\hline
Reference: & what are some ways to help someone with asthma without an inhaler ? \\
\hline
\hline
TRANS (20K) & what happens if a range of a range ofr collides with a range ofr 000r\\
\hline
DDL (20K) & how can i save my without doing waves in a month ?\\
\hline
TRANS (50K) & what can be done to work in a ppr ?\\
\hline
DDL (50K) & what are some of the uses an asthma without an inhaler ?\\
\hline
TRANS (100K) & what is the procedure to be an emergency inr ?\\
\hline
DDL (100K) & how can i help asthma without an inhaler ?\\
\hline
\hline
Source: & how can i get 1 million users to sign up to my app ?\\
\hline
Reference: & how can i get a million users on my social app ? \\
\hline
\hline
TRANS (20K) & how can i get a friend store ?\\
\hline
DDL (20K) & how can i get a game of app on my app ?\\
\hline
TRANS (50K) & how can i get a million million on my startup ?\\
\hline
DDL (50K) & how do i get the first million users app in my app ?\\
\hline
TRANS (100K) & how can i get a million users to write an app ?\\
\hline
DDL (100K) & how do i get a million users to sign up for my app \\
\hline
\hline
Source: & can anyone suggest me the best laptop under 35 k in india ?\\
\hline
Reference: & which is the best laptop under 35 , 000 inr ? \\
\hline
\hline
TRANS (20K) & 	what are the best laptop options available for a laptoprsrsrsrsrsrsrsrsrsrs\\
\hline
DDL (20K) & which is the best laptop to buy in india ?\\
\hline
TRANS (50K) & which is the best laptop under rs . 50000 in india ?\\
\hline
DDL (50K) & what are the best laptops under 35 , 000 in india ?\\
\hline
TRANS (100K) & which is the best laptop to buy under rs . 50000 in india ?\\
\hline
DDL (100K) & which is the best laptop to buy under 35 , 000 in india ?\\
\hline
\hline
Source: & how do i manage my microsoft account ?\\
\hline
Reference: & how can i do manage my microsoft account ? \\
\hline
\hline
TRANS (20K) & how do i manage my google account ?\\
\hline
DDL (20K) & how do i manage my microsoft office ?\\
\hline
TRANS (50K) & how do i manage my ip address ?\\
\hline
DDL (50K) & how do i manage microsoft microsoft office ?\\
\hline
TRANS (100K) & how do i manage my microsoft account ?\\
\hline
DDL (100K) & how do i manage my microsoft account ?\\
\hline
\hline
\end{tabular}}
% \end{adjustwidth}
\end{table}

\begin{table}[h!]
\smallskip
\centering
\caption{{\textbf{Selected paraphrase generation results for semi-supervised model (DDL+VSAR) when incorporating different amounts of unlabelled data (denoted in brackets) and the same amount of labelled data ($20$K), represented in the case of Quora dataset.}}}
\label{tab:semisupervised_qualitative}
\resizebox{.99\linewidth}{!}{
% \begin{adjustwidth}{-2.25in}{0in}{
\smallskip\begin{tabular}{lp{0.70\linewidth}}
\hline
\hline
Source: & is it possible to go to the core of the earth ? \\
\hline
Reference: & if i really wanted to , can i dig all the way to the core of the earth ? \\
\hline
\hline
DDL(20K) & is it possible to go to the earth ?\\
\hline
DDL(20K) + VSAR(20K) & how do i go about the earth ?\\
\hline
DDL(20K) + VSAR(50K) & how do i go about the earth ?\\
\hline
DDL(20K) + VSAR(100K) & is it possible to go to the core of the earth ?\\
\hline		
\hline
Source: & what are best courses for journalism ?\\
\hline
Reference: & what are the best courses on journalism ?\\
\hline
DDL(20K) & what is the best digital marketing course ?\\
\hline
DDL(20K) + VSAR(20K) & what is the best digital marketing agency ?\\
\hline
DDL(20K) + VSAR(50K) & what is the best digital marketing course ?\\
\hline
DDL(20K) + VSAR(100K) & what are the best courses for journalism ?\\
\hline
\hline
Source: & how should i stop thinking about someone ?\\
\hline
Reference: & how do i stop thinking about someone ? \\
\hline
\hline
DDL(20K) & how do i stop thinking about me ?\\
\hline
DDL(20K) + VSAR(20K) & how do i stop thinking about thinking ?\\
\hline
DDL(20K) + VSAR(50K) & how do i stop thinking about something ?\\
\hline
DDL(20K) + VSAR(100K) & how do i stop thinking about someone ?\\
\hline		
\hline
Source: & what motivates all people ?\\
\hline
Reference: & what motivates people ? \\
\hline
\hline
DDL(20K) & why do people often keep all people ?\\
\hline
DDL(20K) + VSAR(20K) & why do people get tattoos ?\\
\hline
DDL(20K) + VSAR(50K) & what inspires to be so hard ?\\
\hline
DDL(20K) + VSAR(100K) & what motivates people ?\\
\hline	
\hline
Source: & how can i get 1 million users to sign up to my app ?\\
\hline
Reference: & how can i get a million users on my social app ? \\
\hline
\hline
DDL(20K) & how can i get a game of app on my app ?\\
\hline
DDL(20K) + VSAR(20K) & how do i get a person from a app ?\\
\hline
DDL(20K) + VSAR(50K) & how can i get a billionaire by youtube ?\\
\hline
DDL(20K) + VSAR(100K) & how can i get 1 million users back from my app ?\\
\hline
\hline
Source: & is vegetarian good for health or non - vegetarian ?\\
\hline
Reference: & which is good food for our health : vegetarian or non - vegetarian ? \\
\hline
\hline
DDL(20K) & is smoking considered a vegetarian vegetarian ?\\
\hline
DDL(20K) + VSAR(20K) & is vegetarian considered good for health ?\\
\hline
DDL(20K) + VSAR(50K) & is vegetarian better than vegetarian ?\\
\hline
DDL(20K) + VSAR(100K) & is vegetarian health good or bad ?\\
\hline						
\hline
Source: & how do i manage my microsoft account ?\\
\hline
Reference: & how can i do manage my microsoft account ? \\
\hline
\hline
DDL(20K) & how do i manage my microsoft office ?\\
\hline
DDL(20K) + VSAR(20K) & how do i manage my microsoft size ?\\
\hline
DDL(20K) + VSAR(50K) & how do i manage my google account ?\\
\hline
DDL(20K) + VSAR(100K) & how do i manage my microsoft account ?\\
\hline	
\hline
\end{tabular}}
% \end{adjustwidth}
% \caption{Selected paraphrase generation results for semi-supervised model (DDL+VSAR) when incorporating different amounts of unlabelled data (denoted in brackets) and the same amount of labelled data ($20$K), represented in the case of Quora dataset.}
% \label{tab:semisupervised_qualitative}
\end{table}

\subsubsection{Learning with Complete Data}

In this section, we present results based on all labelled data in Table \ref{tab: main_quora} and Table \ref{tab: main_mscoco}. Each table comes with three sections. In the first section, we present an upper bound\footnote{Note, here the upper bound does not refer to performance upper bound, rather than an indicator of satisfaction-level performance.} (copying the source as a paraphrase) and a lower bound (randomly selecting {the} ground truth as a paraphrase) calculated based on the test split (as in \cite{chen2020semantically}). This is used as an indication of how well the model performs. In the second section, we present major state-of-the-art models published in recent years. In the third section, we present our own implementation of the Transformer model, which we consider as our absolute baseline, and present results for our models. Our implementation is competitive with the ones reported in recent papers. For our models, DDL is our supervised model, DDL+VSAR is our semi-supervised model, and DDL+VSAR$^{\ast}$ is our model with no prior used. Compared with state-of-the-art supervised models, our models, {in general, achieve statistically significant} better BLEU scores and competitive Rouge scores for both datasets {($p <.05$; Wilcoxon test)}. Our complementary experimental results are presented in Table \ref{tab: complement_quora} and Table \ref{tab: complement_mscoco}, which we compare with two more recent state-of-the-art models. Our models once again achieve {statistically significant} better or competitive performance than the reported {($p <.05$; Wilcoxon test)}, which means our semi-supervised model is competitive with state-of-the-art supervised baselines.

\subsubsection{Qualitative Evaluation for Supervised Learning with labelled Data}
\label{sec: qualitative_supervised_learning}

{In Table} \ref{tab:supervised_qualitative}, {we present examples from the Quora test data set and their corresponding model outputs from our proposed supervised learning model DDL (introduced in section} \ref{sec: ddl}) {and outputs from a very strong baseline model Transformer (denoted as TRANS), using varying amounts of training data. The table first presents the source and golden reference pair, followed by the outputs of the models (DDL and TRANS) trained on $20$K, $50$K, and $100$K labelled dataset pairs. Each example was generated based on a random seed setting of $1000$, ensuring a fair qualitative evaluation; additionally, we always make sure that a smaller amount of labelled pairs is a subset of examples from a larger data size group, this allows us to better quantify the benefits of adding more unlabeled data.} 

{It is quite clear from the results that the generated paraphrase is more accurate in terms of semantic information, and that it matches better with the reference, when more labelled data is used. Although, at the same time, we observe that the advantages of DDL become more significant when the number of labelled pairs is scarce (i.e. with $20$K, the improvement is more significant than with $50$K and $100$K). Additionally, our DDL model demonstrates a clear advantage over the TRANS model in capturing the essence of the information, as seen in its ability to capture critical details (e.g. capture asthma and inhaler in the second example; capture number $35000$ instead of $50000$ in the fourth example). The DDL model also showed more efficient learning, as it was able to achieve comparable results using $50$K data, as opposed to $100$K data for the TRANS model in several examples. These observations further reinforce the effectiveness and efficiency of our proposed DDL model}.

\subsubsection{Qualitative Evaluation for Incorporating Unlabelled Data}

{In table} \ref{tab:semisupervised_qualitative}, {we present qualitative examples from the Quora test data set and their corresponding model outputs from our proposed semi-supervised learning model DDL + VSAR (from section} \ref{sec: vsar-and-ddl}) {given the same amount of labelled data ($20$K, same data instance as in} \ref{sec: qualitative_supervised_learning}) {plus difference size of unlabelled data ($20$K, $50$K and $100$K). For comparison, model output with DDL (section} \ref{sec: ddl}), {trained with the same $20$K examples, is provided. Similarly, as in table} \ref{tab:supervised_qualitative}, {we use the same random seed of $1000$ to generate these examples. In table} \ref{tab:semisupervised_qualitative}, {we first presented the source and golden reference, followed by model outputs trained based on $20$K labelled pairs by the DDL model, the same $20$K pairs used for the DDL+VSAR model (similar to $\text{DDL+VSAR}_{1}$ setting in table} \ref{tab: semi_quora}, \ref{tab: semi_mscoco}, \ref{tab: semi_parabank} and \ref{tab: semi_paramnt}), {the same $20$K with $50$K unlabelled data (extra $30$K) for the DDL+VSAR model and the same $20$K with $100$K unlabelled data (extra $80$K) for the DDL+VSAR model (similar to $\text{DDL+VSAR}_{2}$ setting in table} \ref{tab: semi_quora}, \ref{tab: semi_mscoco}, \ref{tab: semi_parabank} and \ref{tab: semi_paramnt}).

{We can clearly observe that the generated paraphrase matches better with the reference when more unlabelled data is utilised during the training of the DDL+VSAR model. The latent sequence generated when incorporating unlabelled data in the VSAR model improve the paraphrase generation performance. In general, we observe a progressive improvement in capturing the essence of information when incorporating more unlabelled examples. Compared to the preliminary experiment where the VSAR model failed to learn in a fully unsupervised scenario (section} \ref{sec: results_unlabelled_data_only}); {through qualitative results in table} \ref{tab:semisupervised_qualitative} {and quantitative results in table} \ref{tab: semi_quora}, \ref{tab: semi_mscoco}, \ref{tab: semi_parabank} and \ref{tab: semi_paramnt}; {we show that we are able to conduct semi-supervised learning by combining our unsupervised model VSAR and our supervised model DDL.}

\begin{table}[h!]
\smallskip
\centering
\caption{{\textbf{Selected paraphrase generation results for error analysis, represented in the case of Quora dataset.}}}
\label{tab:semisupervised_error_analysis}
\resizebox{.99\linewidth}{!}{
% \begin{adjustwidth}{-2.25in}{0in}{
\smallskip\begin{tabular}{lp{0.70\linewidth}}
\hline
\hline
% Source: & is it possible to go to the core of the earth ? \\
% \hline
% Reference: & if i really wanted to , can i dig all the way to the core of the earth ? \\
% \hline
% \hline
% DDL(20K) & is it possible to go to the earth ?\\
% \hline
% DDL(20K) + VSAR(20K) & how do i go about the earth ?\\
% \hline
% DDL(20K) + VSAR(50K) & how do i go about the earth ?\\
% \hline
% DDL(20K) + VSAR(100K) & is it possible to go to the core of the earth ?\\
% \hline		
% \hline
Source: & what is web marketing ? i want to learn internet marketing ? \\
\hline
Reference: & what ' s the best way to get started in internet marketing ? \\
\hline
% DDL(20K) & what is your purpose of life ?\\
% \hline
DDL(20K) + VSAR(20K) & what is the most valuable website for the internet ?\\
\hline
DDL(20K) + VSAR(50K) & what is the best way to learn internet ?\\
\hline
DDL(20K) + VSAR(100K) & what is the best digital marketing course available online and offline in india and why ?\\
\hline
\hline
Source: & what do you feel is the purpose of life ? \\
\hline
Reference: & from your perspective , what is the purpose of life ? \\
\hline
% DDL(20K) & what is your purpose of life ?\\
% \hline
DDL(20K) + VSAR(20K) & what is the purpose of life ?\\
\hline
DDL(20K) + VSAR(50K) & what is the purpose of your life ?\\
\hline
DDL(20K) + VSAR(100K) & what do you think is the purpose of life ?\\
\hline
\hline
Source: & why do you believe in ghosts and the supernatural ?\\
\hline
Reference: & do you believe in ghosts , why ? \\
\hline
% \hline
% DDL(20K) & how do i stop thinking about me ?\\
% \hline
DDL(20K) + VSAR(20K) & do you believe in ghost ?\\
\hline
DDL(20K) + VSAR(50K) & do you believe in ghosts ?\\
\hline
DDL(20K) + VSAR(100K) & do you think you ' ve ever found in a supernatural event ?\\
\hline		
\hline
Source: & why do so many people post questions on quora that could be easily and thoroughly answered by simply typing the question into any search engine ?\\
\hline
Reference: & why do so many people ask soppy questions on quora , which can easily be googled ? \\
\hline
\hline
% DDL(20K) & why do people often keep all people ?\\
% \hline
DDL(20K) + VSAR(20K) & why do people ask questions on quora instead of googling it ?\\
\hline
DDL(20K) + VSAR(50K) & why do people ask questions on quora instead of googling it ?\\
\hline
DDL(20K) + VSAR(100K) & why do people ask questions on quora that can be answered with a quick web search ?\\
\hline	
\hline
Source: & where can i find delicious cupcakes at gold coast ?\\
\hline
Reference: & where can i found different cupcake flavors in gold coast ? \\
\hline
\hline
% DDL(20K) & how can i get a game of app on my app ?\\
% \hline
DDL(20K) + VSAR(20K) & where can i get best quality gourmet cupcakes in gold coast ?\\
\hline
DDL(20K) + VSAR(50K) & where can i get best quality cupcakes in gold coast ?\\
\hline
DDL(20K) + VSAR(100K) & where can i get best flavors , designs and decorations for cupcakes at gold coast ?\\
\hline
\hline
Source: & is vegetarian good for health or non - vegetarian ?\\
\hline
Reference: & which is good food for our health : vegetarian or non - vegetarian ? \\
\hline
\hline
DDL(20K) & is smoking considered a vegetarian vegetarian ?\\
\hline
DDL(20K) + VSAR(20K) & is vegetarian considered good for health ?\\
\hline
DDL(20K) + VSAR(50K) & is vegetarian better than vegetarian ?\\
\hline
DDL(20K) + VSAR(100K) & is vegetarian health good or bad ?\\
\hline						
\hline
% Source: & how do i manage my microsoft account ?\\
% \hline
% Reference: & how can i do manage my microsoft account ? \\
% \hline
% \hline
% DDL(20K) & how do i manage my microsoft office ?\\
% \hline
% DDL(20K) + VSAR(20K) & how do i manage my microsoft size ?\\
% \hline
% DDL(20K) + VSAR(50K) & how do i manage my google account ?\\
% \hline
% DDL(20K) + VSAR(100K) & how do i manage my microsoft account ?\\
% \hline	
% \hline
\end{tabular}}
% \end{adjustwidth}
% \caption{Selected paraphrase generation results for semi-supervised model (DDL+VSAR) when incorporating different amounts of unlabelled data (denoted in brackets) and the same amount of labelled data ($20$K), represented in the case of Quora dataset.}
% \label{tab:semisupervised_qualitative}
\end{table}

\subsubsection{Error Analysis}

{In this section, we present an error analysis of our proposed model. In general, our model works well for semi-supervised settings, as demonstrated in Tables} \ref{tab: semi_quora},\ref{tab: semi_mscoco}, \ref{tab: semi_parabank} and \ref{tab: semi_paramnt}, {however, there are cases that the generated examples from our models get worth quality or no improvements in terms of quality when more unlabeled pairs are utilised. We find that in cases where the DDL+VSAR model has no improvement when seeing more unlabelled data, the golden reference of test examples is relatively easy to paraphrase and is presented in shorter sentence form. However, we also observed scenarios when our DDL+VSAR model overcomplex paraphrases when more unlabelled data points are used, as shown in table} \ref{tab:semisupervised_error_analysis}. {We notice that observing unlabelled data increase the model's ability to handle more complex paraphrase, however, at the same time encourage the model to return a complex answer. This is an interesting discovery and suggests that the quality of unlabelled data also plays an important role in the semi-supervised learning process. }

% the most common sources of errors are due to mislabeling of data, errors in the model's training data, and errors in the model's implementation. We also find that the model's performance is affected by the amount of data available for training and the complexity of the model.

% We suggest a number of ways to improve the model's performance, including using more data for training, using more complex models, and using more accurate labeling of data. We also suggest a number of ways to improve the model's interpretability, including using more visualization tools and using more explanations of the model's decisions.

\subsubsection{Algorithm Run Time}

{In this section, we further discuss the algorithm run time and GPU memory requirement of our proposed models. Regarding the DDL model, it is equivalent to training a standard transformer model with additional cost terms (from two directions: source to target and target to source). Hence, training a DDL model does not require additional run time and requires no additional GPU memory. Regarding the VSAR model alone, since we are sharing the parameters for source reconstruction and latent inference (as shown in Figure} \ref{fig: VSAR_Model}), {the cost of training is similar to the DDL model, plus some additional GPU memory cost due to the need of saving gradient from latent samples in order to perform back-propagation. Since a single sample is used during our training, the extra GPU memory cost is not significant, and we do not recognise the extra run time for training the VSAR model. Regarding the semi-supervised learning model, DDL+VSAR, we need to first pre-train a DDL model from the labelled data, then initialise the model weight from the DDL model and jointly fine-tune the DDL+VSAR model from both labelled and unlabelled data. The total run time is doubled compared to the process of pre-training and fine-tuning. Although we find that pre-training already set the model at a near-optimal parameter space, and hence we could potentially use fewer epochs for the semi-supervised learning to ensure the algorithm scales well to large-scale data.}

\section{Limitations and Future Work}

{Here we briefly discuss two main limitations which are identified in this research with each associated future work. The first limitation involves the quantitative evaluation metrics, such as BLEU and Rouge score, used in this research. These metrics are based on the overlap of n-gram contexts between generated outputs and reference text, and while they are commonly used to compare with published results (as shown in Table} \ref{tab: main_quora}, \ref{tab: main_mscoco}, \ref{tab: complement_quora} and \ref{tab: complement_mscoco}) {, they cannot directly assess the quality of the generated text. More recent evaluation metrics, such as BERTscore} \cite{zhangbertscore}, {have been proposed, but there is still no universal agreement on what are the best quantitative measurements. One future work direction is to look into this area and propose better evaluation metrics for capturing the quality of generated paraphrases, which will ultimately create a meaningful impact on natural language generation research.}

{The second limitation is the absence of user-based studies to evaluate systems implemented based on our proposed methods, which are often considered a more comprehensive measure of model performance. Although in this paper, we observed promising results in section} \ref{sec: results_and_discussion}. {Current evaluations in this work rely on public benchmark datasets, which may be biased towards the subset of data the model is trained on. Due to resource constraints, user-based evaluations were not performed in this research and represented a promising area for future study.}

% {For future work, in this paper, we have explored a more structured approach to modelling unobserved paraphrase as a discrete latent variable in the graphical model. A future direction worth exploring is to introduce hierarchy in latent variable representations in this graphical model. One possible direction is to explore efficient spectral clustering algorithms} \cite{li2018rank,li2018dynamic}. {This will potentially allow us to produce diverse paraphrases, each from local clusters (as opposed to topic information) and allow the possibility of finer-grained control for topic-based paraphrase generation.}

{For future work, in this paper, we have explored a more structured approach to modelling unobserved paraphrase as a discrete latent variable. A future direction worth exploring is to apply and extend this technique in pre-trained language models (PLMs). Recent advancements in PLMs suggest that they can achieve very good performance given a relatively small proportion of data or even perform well in few-shot or zero-shot scenarios in document retrieval tasks. However, challenges still exist in the cases of natural language generation (NLG) tasks, and our proposed method in this paper can serve as a stepping stone to using both labelled and unlabelled data for NLG tasks.

% suggest we introduce hierarchy in latent variable representations in this graphical model. One possible direction is to explore efficient spectral clustering algorithms} \cite{li2018rank,li2018dynamic}. {This will potentially allow us to produce diverse paraphrases, each from local clusters (as opposed to topic information) and allow the possibility of finer-grained control for topic-based paraphrase generation.}

\section{Conclusions}

In this paper, we have introduced a semi-supervised deep generative model for paraphrase generation. The unsupervised model (VSAR) is based on the variational auto-encoding framework and provides an effective method to handle missing labels. The supervised model (DDL) conducts dual learning and injects supervised information into the unsupervised model. With our novel knowledge-reinforced-learning {(KRL)} scheme, we empirically demonstrate that semi-supervised learning benefits our combined model, given unlabelled data and a fraction of the paired data. The evaluation results show that our combined model improves upon a very strong baseline model in a semi-supervised setting. We also observe that, even for the full dataset, our combined model achieves competitive performance with the state-of-the-art models for two paraphrase generation benchmark datasets. Additionally, we are able to model language as a discrete latent variable sequence for paraphrase generation tasks. Importantly, the resultant generative model is able to {exploit both supervised and unsupervised data in sequence-to-sequence tasks effectively}.

% \begin{table}[h!]
% \smallskip
% \centering
% \caption{Complement Results for MSCOCO.}
% \resizebox{0.95\columnwidth}{!}{
% \smallskip\begin{tabular}{lccc}
% \hline
% Model & B-4 & self-B & i-B \\
% % \hline
% % Upper Bound (Copy Source) & 63.36 & 49.99 & 40.47  \\
% % Lower Bound (Random Select)& 16.10 & 4.50 & 1.94 \\
% \hline
% Separator \cite{hosking2021factorising}  & $20.59$ & $12.76$ & $13.92$ \\
% HRQ-VAE \cite{hosking2022hierarchical}  & $27.90$ & $16.58$ & $19.04$ \\
% \hline
% Transformer (our implementation) & $26.87$ & $\textbf{13.50}$ & $18.79$ \\
% % DDL (our model) & $27.88\uparrow$ & $14.83\downarrow$ & $\textbf{19.34}\uparrow$ \\
% DDL + SVAR (our model) & $27.87\uparrow$ & $15.42\downarrow$ & $19.21\uparrow$ \\
% DDL + SVAR$^{\ast}$ (our model) & $\textbf{27.92}\uparrow$ & $15.21\downarrow$ & $\textbf{19.29}\uparrow$ \\
% \hline
% \end{tabular}}
% \label{tab: complement_mscoco}
% \end{table}

% \appendix
% \section{My Appendix}
% Appendix sections are coded under \verb+\appendix+.

% \verb+\printcredits+ command is used after appendix sections to list 
% author credit taxonomy contribution roles tagged using \verb+\credit+ 
% in frontmatter.


% \bibliographystyle{cas-model2-names}

\bibliographystyle{unsrt}  
\bibliography{references}

\begin{thebibliography}{10}

\bibitem{dong2017learning}
Li~Dong, Jonathan Mallinson, Siva Reddy, and Mirella Lapata.
\newblock Learning to paraphrase for question answering.
\newblock In {\em Proceedings of the 2017 Conference on Empirical Methods in
  Natural Language Processing}, pages 875--886, Copenhagen, Denmark, 2017.
  Association for Computational Linguistics.

\bibitem{lee2006beyond}
Minsuk Lee, James Cimino, Hai~Ran Zhu, Carl Sable, Vijay Shanker, John Ely, and
  Hong Yu.
\newblock Beyond information retrieval—medical question answering.
\newblock In {\em AMIA annual symposium proceedings}, volume 2006, page 469.
  American Medical Informatics Association, 2006.

\bibitem{yao2014information}
Xuchen Yao and Benjamin Van~Durme.
\newblock Information extraction over structured data: Question answering with
  {F}reebase.
\newblock In {\em Proceedings of the 52nd Annual Meeting of the Association for
  Computational Linguistics (Volume 1: Long Papers)}, pages 956--966,
  Baltimore, Maryland, 2014. Association for Computational Linguistics.

\bibitem{liu2008understanding}
Yuanjie Liu, Shasha Li, Yunbo Cao, Chin-Yew Lin, Dingyi Han, and Yong Yu.
\newblock Understanding and summarizing answers in community-based question
  answering services.
\newblock In {\em Proceedings of the 22nd International Conference on
  Computational Linguistics (Coling 2008)}, pages 497--504, Manchester, UK,
  2008. Coling 2008 Organizing Committee.

\bibitem{miller2019explanation}
Tim Miller.
\newblock Explanation in artificial intelligence: Insights from the social
  sciences.
\newblock {\em Artificial intelligence}, 267:1--38, 2019.

\bibitem{hosking2022hierarchical}
Tom Hosking, Hao Tang, and Mirella Lapata.
\newblock Hierarchical sketch induction for paraphrase generation.
\newblock {\em arXiv preprint arXiv:2203.03463}, 2022.

\bibitem{kauchak2006paraphrasing}
David Kauchak and Regina Barzilay.
\newblock Paraphrasing for automatic evaluation.
\newblock In {\em Proceedings of the Human Language Technology Conference of
  the {NAACL}, Main Conference}, pages 455--462, New York City, USA, 2006.
  Association for Computational Linguistics.

\bibitem{narayan2016paraphrase}
Shashi Narayan, Siva Reddy, and Shay~B. Cohen.
\newblock Paraphrase generation from latent-variable {PCFG}s for semantic
  parsing.
\newblock In {\em Proceedings of the 9th International Natural Language
  Generation conference}, pages 153--162, Edinburgh, UK, 2016. Association for
  Computational Linguistics.

\bibitem{miller1995wordnet}
George~A. Miller.
\newblock {W}ord{N}et: A lexical database for {E}nglish.
\newblock In {\em Speech and Natural Language: Proceedings of a Workshop Held
  at Harriman, New York, {F}ebruary 23-26, 1992}, 1992.

\bibitem{kumar2020syntax}
Ashutosh Kumar, Kabir Ahuja, Raghuram Vadapalli, and Partha Talukdar.
\newblock Syntax-guided controlled generation of paraphrases.
\newblock {\em Transactions of the Association for Computational Linguistics},
  8:329--345, 2020.

\bibitem{zhou2021paraphrase}
Jianing Zhou and Suma Bhat.
\newblock Paraphrase generation: A survey of the state of the art.
\newblock In {\em Proceedings of the 2021 Conference on Empirical Methods in
  Natural Language Processing}, pages 5075--5086, 2021.

\bibitem{meng2021conrpg}
Yuxian Meng, Xiang Ao, Qing He, Xiaofei Sun, Qinghong Han, Fei Wu, Jiwei Li,
  et~al.
\newblock Conrpg: Paraphrase generation using contexts as regularizer.
\newblock {\em arXiv preprint arXiv:2109.00363}, 2021.

\bibitem{su2021keep}
Yixuan Su, David Vandyke, Simon Baker, Yan Wang, and Nigel Collier.
\newblock Keep the primary, rewrite the secondary: A two-stage approach for
  paraphrase generation.
\newblock In {\em Findings of the Association for Computational Linguistics:
  ACL-IJCNLP 2021}, pages 560--569, Online, 2021. Association for Computational
  Linguistics.

\bibitem{sutskever2014sequence}
Ilya Sutskever, Oriol Vinyals, and Quoc~V. Le.
\newblock Sequence to sequence learning with neural networks.
\newblock In Zoubin Ghahramani, Max Welling, Corinna Cortes, Neil~D. Lawrence,
  and Kilian~Q. Weinberger, editors, {\em Advances in Neural Information
  Processing Systems 27: Annual Conference on Neural Information Processing
  Systems 2014, December 8-13 2014, Montreal, Quebec, Canada}, pages
  3104--3112, 2014.

\bibitem{zhu2005semi}
Xiaojin~Jerry Zhu.
\newblock Semi-supervised learning literature survey.
\newblock 2005.

\bibitem{van2020survey}
Jesper~E Van~Engelen and Holger~H Hoos.
\newblock A survey on semi-supervised learning.
\newblock {\em Machine learning}, 109(2):373--440, 2020.

\bibitem{kingma2013auto}
Diederik~P. Kingma and Max Welling.
\newblock Auto-encoding variational bayes.
\newblock In Yoshua Bengio and Yann LeCun, editors, {\em 2nd International
  Conference on Learning Representations, {ICLR} 2014, Banff, AB, Canada, April
  14-16, 2014, Conference Track Proceedings}, 2014.

\bibitem{rezende2014stochastic}
Danilo~Jimenez Rezende, Shakir Mohamed, and Daan Wierstra.
\newblock Stochastic backpropagation and approximate inference in deep
  generative models.
\newblock In {\em Proceedings of the 31th International Conference on Machine
  Learning, {ICML} 2014, Beijing, China, 21-26 June 2014}, volume~32 of {\em
  {JMLR} Workshop and Conference Proceedings}, pages 1278--1286. JMLR.org,
  2014.

\bibitem{mnih2014neural}
Andriy Mnih and Karol Gregor.
\newblock Neural variational inference and learning in belief networks.
\newblock In {\em Proceedings of the 31th International Conference on Machine
  Learning, {ICML} 2014, Beijing, China, 21-26 June 2014}, volume~32 of {\em
  {JMLR} Workshop and Conference Proceedings}, pages 1791--1799. JMLR.org,
  2014.

\bibitem{hinton2006reducing}
Geoffrey~E Hinton and Ruslan~R Salakhutdinov.
\newblock Reducing the dimensionality of data with neural networks.
\newblock {\em science}, 313(5786):504--507, 2006.

\bibitem{fu2020paraphrase}
Yao Fu, Yansong Feng, and John~P. Cunningham.
\newblock Paraphrase generation with latent bag of words.
\newblock In Hanna~M. Wallach, Hugo Larochelle, Alina Beygelzimer, Florence
  d'Alch{\'{e}}{-}Buc, Emily~B. Fox, and Roman Garnett, editors, {\em Advances
  in Neural Information Processing Systems 32: Annual Conference on Neural
  Information Processing Systems 2019, NeurIPS 2019, December 8-14, 2019,
  Vancouver, BC, Canada}, pages 13623--13634, 2019.

\bibitem{hosking2021factorising}
Tom Hosking and Mirella Lapata.
\newblock Factorising meaning and form for intent-preserving paraphrasing.
\newblock In {\em Proceedings of the 59th Annual Meeting of the Association for
  Computational Linguistics and the 11th International Joint Conference on
  Natural Language Processing (Volume 1: Long Papers)}, pages 1405--1418,
  Online, 2021. Association for Computational Linguistics.

\bibitem{he2016dual}
Di~He, Yingce Xia, Tao Qin, Liwei Wang, Nenghai Yu, Tie{-}Yan Liu, and
  Wei{-}Ying Ma.
\newblock Dual learning for machine translation.
\newblock In Daniel~D. Lee, Masashi Sugiyama, Ulrike von Luxburg, Isabelle
  Guyon, and Roman Garnett, editors, {\em Advances in Neural Information
  Processing Systems 29: Annual Conference on Neural Information Processing
  Systems 2016, December 5-10, 2016, Barcelona, Spain}, pages 820--828, 2016.

\bibitem{su2019dual}
Shang-Yu Su, Chao-Wei Huang, and Yun-Nung Chen.
\newblock Dual supervised learning for natural language understanding and
  generation.
\newblock In {\em Proceedings of the 57th Annual Meeting of the Association for
  Computational Linguistics}, pages 5472--5477, Florence, Italy, 2019.
  Association for Computational Linguistics.

\bibitem{su2020dual}
Shang-Yu Su, Yung-Sung Chuang, and Yun-Nung Chen.
\newblock Dual inference for improving language understanding and generation.
\newblock In {\em Findings of the Association for Computational Linguistics:
  EMNLP 2020}, pages 4930--4936, Online, 2020. Association for Computational
  Linguistics.

\bibitem{su2020towards}
Shang-Yu Su, Chao-Wei Huang, and Yun-Nung Chen.
\newblock Towards unsupervised language understanding and generation by joint
  dual learning.
\newblock In {\em Proceedings of the 58th Annual Meeting of the Association for
  Computational Linguistics}, pages 671--680, Online, 2020. Association for
  Computational Linguistics.

\bibitem{pavlick2015ppdb}
Ellie Pavlick, Pushpendre Rastogi, Juri Ganitkevitch, Benjamin Van~Durme, and
  Chris Callison-Burch.
\newblock {PPDB} 2.0: Better paraphrase ranking, fine-grained entailment
  relations, word embeddings, and style classification.
\newblock In {\em Proceedings of the 53rd Annual Meeting of the Association for
  Computational Linguistics and the 7th International Joint Conference on
  Natural Language Processing (Volume 2: Short Papers)}, pages 425--430,
  Beijing, China, 2015. Association for Computational Linguistics.

\bibitem{gupta2018deep}
Ankush Gupta, Arvind Agarwal, Prawaan Singh, and Piyush Rai.
\newblock A deep generative framework for paraphrase generation.
\newblock In Sheila~A. McIlraith and Kilian~Q. Weinberger, editors, {\em
  Proceedings of the Thirty-Second {AAAI} Conference on Artificial
  Intelligence, (AAAI-18), the 30th innovative Applications of Artificial
  Intelligence (IAAI-18), and the 8th {AAAI} Symposium on Educational Advances
  in Artificial Intelligence (EAAI-18), New Orleans, Louisiana, USA, February
  2-7, 2018}, pages 5149--5156. {AAAI} Press, 2018.

\bibitem{chen2022mcpg}
Yi~Chen, Haiyun Jiang, Lemao Liu, Rui Wang, Shuming Shi, and Ruifeng Xu.
\newblock Mcpg: A flexible multi-level controllable framework for unsupervised
  paraphrase generation.
\newblock In {\em Findings of the Association for Computational Linguistics:
  EMNLP 2022}, pages 5948--5958, 2022.

\bibitem{xie2023visual}
Jiayuan Xie, Jiali Chen, Yi~Cai, Qingbao Huang, and Qing Li.
\newblock Visual paraphrase generation with key information retained.
\newblock {\em ACM Transactions on Multimedia Computing, Communications and
  Applications}, 2023.

\bibitem{martins-etal-2019-latent}
Andr{\'e} F.~T. Martins, Tsvetomila Mihaylova, Nikita Nangia, and Vlad Niculae.
\newblock Latent structure models for natural language processing.
\newblock In {\em Proceedings of the 57th Annual Meeting of the Association for
  Computational Linguistics: Tutorial Abstracts}, pages 1--5, Florence, Italy,
  July 2019. Association for Computational Linguistics.

\bibitem{miao2016language}
Yishu Miao and Phil Blunsom.
\newblock Language as a latent variable: Discrete generative models for
  sentence compression.
\newblock In {\em Proceedings of the 2016 Conference on Empirical Methods in
  Natural Language Processing}, pages 319--328, Austin, Texas, 2016.
  Association for Computational Linguistics.

\bibitem{miao2016neural}
Yishu Miao, Lei Yu, and Phil Blunsom.
\newblock Neural variational inference for text processing.
\newblock In Maria{-}Florina Balcan and Kilian~Q. Weinberger, editors, {\em
  Proceedings of the 33nd International Conference on Machine Learning, {ICML}
  2016, New York City, NY, USA, June 19-24, 2016}, volume~48 of {\em {JMLR}
  Workshop and Conference Proceedings}, pages 1727--1736. JMLR.org, 2016.

\bibitem{kim2018tutorial}
Yoon Kim, Sam Wiseman, and Alexander~M Rush.
\newblock A tutorial on deep latent variable models of natural language.
\newblock {\em arXiv preprint arXiv:1812.06834}, 2018.

\bibitem{bowman2015generating}
Samuel~R. Bowman, Luke Vilnis, Oriol Vinyals, Andrew Dai, Rafal Jozefowicz, and
  Samy Bengio.
\newblock Generating sentences from a continuous space.
\newblock In {\em Proceedings of The 20th {SIGNLL} Conference on Computational
  Natural Language Learning}, pages 10--21, Berlin, Germany, 2016. Association
  for Computational Linguistics.

\bibitem{dieng2019avoiding}
Adji~B. Dieng, Yoon Kim, Alexander~M. Rush, and David~M. Blei.
\newblock Avoiding latent variable collapse with generative skip models.
\newblock In Kamalika Chaudhuri and Masashi Sugiyama, editors, {\em The 22nd
  International Conference on Artificial Intelligence and Statistics, {AISTATS}
  2019, 16-18 April 2019, Naha, Okinawa, Japan}, volume~89 of {\em Proceedings
  of Machine Learning Research}, pages 2397--2405. {PMLR}, 2019.

\bibitem{he2019lagging}
Junxian He, Daniel Spokoyny, Graham Neubig, and Taylor Berg{-}Kirkpatrick.
\newblock Lagging inference networks and posterior collapse in variational
  autoencoders.
\newblock In {\em 7th International Conference on Learning Representations,
  {ICLR} 2019, New Orleans, LA, USA, May 6-9, 2019}. OpenReview.net, 2019.

\bibitem{higgins2016beta}
Irina Higgins, Lo{\"{\i}}c Matthey, Arka Pal, Christopher Burgess, Xavier
  Glorot, Matthew Botvinick, Shakir Mohamed, and Alexander Lerchner.
\newblock beta-vae: Learning basic visual concepts with a constrained
  variational framework.
\newblock In {\em 5th International Conference on Learning Representations,
  {ICLR} 2017, Toulon, France, April 24-26, 2017, Conference Track
  Proceedings}. OpenReview.net, 2017.

\bibitem{razavi2019preventing}
Ali Razavi, A{\"{a}}ron van~den Oord, Ben Poole, and Oriol Vinyals.
\newblock Preventing posterior collapse with delta-vaes.
\newblock In {\em 7th International Conference on Learning Representations,
  {ICLR} 2019, New Orleans, LA, USA, May 6-9, 2019}. OpenReview.net, 2019.

\bibitem{wang2021posterior}
Yixin Wang, David Blei, and John~P Cunningham.
\newblock Posterior collapse and latent variable non-identifiability.
\newblock {\em Advances in Neural Information Processing Systems}, 34, 2021.

\bibitem{wen2017latent}
Tsung{-}Hsien Wen, Yishu Miao, Phil Blunsom, and Steve~J. Young.
\newblock Latent intention dialogue models.
\newblock In Doina Precup and Yee~Whye Teh, editors, {\em Proceedings of the
  34th International Conference on Machine Learning, {ICML} 2017, Sydney, NSW,
  Australia, 6-11 August 2017}, volume~70 of {\em Proceedings of Machine
  Learning Research}, pages 3732--3741. {PMLR}, 2017.

\bibitem{mnih2014recurrent}
Volodymyr Mnih, Nicolas Heess, Alex Graves, and Koray Kavukcuoglu.
\newblock Recurrent models of visual attention.
\newblock In Zoubin Ghahramani, Max Welling, Corinna Cortes, Neil~D. Lawrence,
  and Kilian~Q. Weinberger, editors, {\em Advances in Neural Information
  Processing Systems 27: Annual Conference on Neural Information Processing
  Systems 2014, December 8-13 2014, Montreal, Quebec, Canada}, pages
  2204--2212, 2014.

\bibitem{jang2016categorical}
Eric Jang, Shixiang Gu, and Ben Poole.
\newblock Categorical reparameterization with gumbel-softmax.
\newblock In {\em 5th International Conference on Learning Representations,
  {ICLR} 2017, Toulon, France, April 24-26, 2017, Conference Track
  Proceedings}. OpenReview.net, 2017.

\bibitem{maddison2016concrete}
Chris~J. Maddison, Andriy Mnih, and Yee~Whye Teh.
\newblock The concrete distribution: {A} continuous relaxation of discrete
  random variables.
\newblock In {\em 5th International Conference on Learning Representations,
  {ICLR} 2017, Toulon, France, April 24-26, 2017, Conference Track
  Proceedings}. OpenReview.net, 2017.

\bibitem{choi2018learning}
Jihun Choi, Kang~Min Yoo, and Sang{-}goo Lee.
\newblock Learning to compose task-specific tree structures.
\newblock In Sheila~A. McIlraith and Kilian~Q. Weinberger, editors, {\em
  Proceedings of the Thirty-Second {AAAI} Conference on Artificial
  Intelligence, (AAAI-18), the 30th innovative Applications of Artificial
  Intelligence (IAAI-18), and the 8th {AAAI} Symposium on Educational Advances
  in Artificial Intelligence (EAAI-18), New Orleans, Louisiana, USA, February
  2-7, 2018}, pages 5094--5101. {AAAI} Press, 2018.

\bibitem{he2020probabilistic}
Junxian He, Xinyi Wang, Graham Neubig, and Taylor Berg{-}Kirkpatrick.
\newblock A probabilistic formulation of unsupervised text style transfer.
\newblock In {\em 8th International Conference on Learning Representations,
  {ICLR} 2020, Addis Ababa, Ethiopia, April 26-30, 2020}. OpenReview.net, 2020.

\bibitem{mercatali2021disentangling}
Giangiacomo Mercatali and Andr{\'e} Freitas.
\newblock Disentangling generative factors in natural language with discrete
  variational autoencoders.
\newblock {\em arXiv preprint arXiv:2109.07169}, 2021.

\bibitem{xie2019reparameterizable}
Sang~Michael Xie and Stefano Ermon.
\newblock Reparameterizable subset sampling via continuous relaxations.
\newblock In Sarit Kraus, editor, {\em Proceedings of the Twenty-Eighth
  International Joint Conference on Artificial Intelligence, {IJCAI} 2019,
  Macao, China, August 10-16, 2019}, pages 3919--3925. ijcai.org, 2019.

\bibitem{niculae2023discrete}
Vlad Niculae, Caio~F Corro, Nikita Nangia, Tsvetomila Mihaylova, and
  Andr{\'e}~FT Martins.
\newblock Discrete latent structure in neural networks.
\newblock {\em arXiv preprint arXiv:2301.07473}, 2023.

\bibitem{huang2023make}
Rongjie Huang, Jiawei Huang, Dongchao Yang, Yi~Ren, Luping Liu, Mingze Li,
  Zhenhui Ye, Jinglin Liu, Xiang Yin, and Zhou Zhao.
\newblock Make-an-audio: Text-to-audio generation with prompt-enhanced
  diffusion models.
\newblock {\em arXiv preprint arXiv:2301.12661}, 2023.

\bibitem{schneider2023mo}
Flavio Schneider, Zhijing Jin, and Bernhard Sch{\"o}lkopf.
\newblock Mo$\backslash$\^{} usai: Text-to-music generation with long-context
  latent diffusion.
\newblock {\em arXiv preprint arXiv:2301.11757}, 2023.

\bibitem{luong2015multi}
Minh{-}Thang Luong, Quoc~V. Le, Ilya Sutskever, Oriol Vinyals, and Lukasz
  Kaiser.
\newblock Multi-task sequence to sequence learning.
\newblock In Yoshua Bengio and Yann LeCun, editors, {\em 4th International
  Conference on Learning Representations, {ICLR} 2016, San Juan, Puerto Rico,
  May 2-4, 2016, Conference Track Proceedings}, 2016.

\bibitem{guo2018dynamic}
Han Guo, Ramakanth Pasunuru, and Mohit Bansal.
\newblock Dynamic multi-level multi-task learning for sentence simplification.
\newblock In {\em Proceedings of the 27th International Conference on
  Computational Linguistics}, pages 462--476, Santa Fe, New Mexico, USA, 2018.
  Association for Computational Linguistics.

\bibitem{guo2018soft}
Han Guo, Ramakanth Pasunuru, and Mohit Bansal.
\newblock Soft layer-specific multi-task summarization with entailment and
  question generation.
\newblock In {\em Proceedings of the 56th Annual Meeting of the Association for
  Computational Linguistics (Volume 1: Long Papers)}, pages 687--697,
  Melbourne, Australia, 2018. Association for Computational Linguistics.

\bibitem{wang2020multi}
Yiren Wang, ChengXiang Zhai, and Hany Hassan.
\newblock Multi-task learning for multilingual neural machine translation.
\newblock In {\em Proceedings of the 2020 Conference on Empirical Methods in
  Natural Language Processing (EMNLP)}, pages 1022--1034, Online, 2020.
  Association for Computational Linguistics.

\bibitem{du2020self}
Jingfei Du, Edouard Grave, Beliz Gunel, Vishrav Chaudhary, Onur Celebi, Michael
  Auli, Veselin Stoyanov, and Alexis Conneau.
\newblock Self-training improves pre-training for natural language
  understanding.
\newblock In {\em Proceedings of the 2021 Conference of the North American
  Chapter of the Association for Computational Linguistics: Human Language
  Technologies}, pages 5408--5418, Online, 2021. Association for Computational
  Linguistics.

\bibitem{chang2021jointly}
Ernie Chang, Vera Demberg, and Alex Marin.
\newblock Jointly improving language understanding and generation with
  quality-weighted weak supervision of automatic labeling.
\newblock In {\em Proceedings of the 16th Conference of the European Chapter of
  the Association for Computational Linguistics: Main Volume}, pages 818--829,
  Online, 2021. Association for Computational Linguistics.

\bibitem{williams1989learning}
Ronald~J Williams and David Zipser.
\newblock A learning algorithm for continually running fully recurrent neural
  networks.
\newblock {\em Neural computation}, 1(2):270--280, 1989.

\bibitem{burrows2013paraphrase}
Steven Burrows, Martin Potthast, and Benno Stein.
\newblock Paraphrase acquisition via crowdsourcing and machine learning.
\newblock {\em ACM Transactions on Intelligent Systems and Technology (TIST)},
  4(3):1--21, 2013.

\bibitem{cao2017joint}
Ziqiang Cao, Chuwei Luo, Wenjie Li, and Sujian Li.
\newblock Joint copying and restricted generation for paraphrase.
\newblock In Satinder~P. Singh and Shaul Markovitch, editors, {\em Proceedings
  of the Thirty-First {AAAI} Conference on Artificial Intelligence, February
  4-9, 2017, San Francisco, California, {USA}}, pages 3152--3158. {AAAI} Press,
  2017.

\bibitem{chen2018learning}
Jianbo Chen, Le~Song, Martin~J. Wainwright, and Michael~I. Jordan.
\newblock Learning to explain: An information-theoretic perspective on model
  interpretation.
\newblock In Jennifer~G. Dy and Andreas Krause, editors, {\em Proceedings of
  the 35th International Conference on Machine Learning, {ICML} 2018,
  Stockholmsm{\"{a}}ssan, Stockholm, Sweden, July 10-15, 2018}, volume~80 of
  {\em Proceedings of Machine Learning Research}, pages 882--891. {PMLR}, 2018.

\bibitem{balin2019concrete}
Muhammed~Fatih Balin, Abubakar Abid, and James~Y. Zou.
\newblock Concrete autoencoders: Differentiable feature selection and
  reconstruction.
\newblock In Kamalika Chaudhuri and Ruslan Salakhutdinov, editors, {\em
  Proceedings of the 36th International Conference on Machine Learning, {ICML}
  2019, 9-15 June 2019, Long Beach, California, {USA}}, volume~97 of {\em
  Proceedings of Machine Learning Research}, pages 444--453. {PMLR}, 2019.

\bibitem{bengio2013estimating}
Yoshua Bengio, Nicholas L{\'e}onard, and Aaron Courville.
\newblock Estimating or propagating gradients through stochastic neurons for
  conditional computation.
\newblock {\em arXiv preprint arXiv:1308.3432}, 2013.

\bibitem{vaswani2017attention}
Ashish Vaswani, Noam Shazeer, Niki Parmar, Jakob Uszkoreit, Llion Jones,
  Aidan~N. Gomez, Lukasz Kaiser, and Illia Polosukhin.
\newblock Attention is all you need.
\newblock In Isabelle Guyon, Ulrike von Luxburg, Samy Bengio, Hanna~M. Wallach,
  Rob Fergus, S.~V.~N. Vishwanathan, and Roman Garnett, editors, {\em Advances
  in Neural Information Processing Systems 30: Annual Conference on Neural
  Information Processing Systems 2017, December 4-9, 2017, Long Beach, CA,
  {USA}}, pages 5998--6008, 2017.

\bibitem{yang2018unsupervised}
Zichao Yang, Zhiting Hu, Chris Dyer, Eric~P. Xing, and Taylor
  Berg{-}Kirkpatrick.
\newblock Unsupervised text style transfer using language models as
  discriminators.
\newblock In Samy Bengio, Hanna~M. Wallach, Hugo Larochelle, Kristen Grauman,
  Nicol{\`{o}} Cesa{-}Bianchi, and Roman Garnett, editors, {\em Advances in
  Neural Information Processing Systems 31: Annual Conference on Neural
  Information Processing Systems 2018, NeurIPS 2018, December 3-8, 2018,
  Montr{\'{e}}al, Canada}, pages 7298--7309, 2018.

\bibitem{lin2014microsoft}
Tsung-Yi Lin, Michael Maire, Serge Belongie, James Hays, Pietro Perona, Deva
  Ramanan, Piotr Doll{\'a}r, and C~Lawrence Zitnick.
\newblock Microsoft coco: Common objects in context.
\newblock In {\em European conference on computer vision}, pages 740--755.
  Springer, 2014.

\bibitem{devlin2018bert}
Jacob Devlin, Ming-Wei Chang, Kenton Lee, and Kristina Toutanova.
\newblock {BERT}: Pre-training of deep bidirectional transformers for language
  understanding.
\newblock In {\em Proceedings of the 2019 Conference of the North {A}merican
  Chapter of the Association for Computational Linguistics: Human Language
  Technologies, Volume 1 (Long and Short Papers)}, pages 4171--4186,
  Minneapolis, Minnesota, 2019. Association for Computational Linguistics.

\bibitem{li2019decomposable}
Zichao Li, Xin Jiang, Lifeng Shang, and Qun Liu.
\newblock Decomposable neural paraphrase generation.
\newblock In {\em Proceedings of the 57th Annual Meeting of the Association for
  Computational Linguistics}, pages 3403--3414, Florence, Italy, 2019.
  Association for Computational Linguistics.

\bibitem{hu2019parabank}
J~Edward Hu, Rachel Rudinger, Matt Post, and Benjamin Van~Durme.
\newblock Parabank: Monolingual bitext generation and sentential paraphrasing
  via lexically-constrained neural machine translation.
\newblock In {\em Proceedings of the AAAI Conference on Artificial
  Intelligence}, volume~33, pages 6521--6528, 2019.

\bibitem{wieting2017paranmt}
John Wieting and Kevin Gimpel.
\newblock Paranmt-50m: Pushing the limits of paraphrastic sentence embeddings
  with millions of machine translations.
\newblock {\em arXiv preprint arXiv:1711.05732}, 2017.

\bibitem{pennington2014glove}
Jeffrey Pennington, Richard Socher, and Christopher Manning.
\newblock {G}lo{V}e: Global vectors for word representation.
\newblock In {\em Proceedings of the 2014 Conference on Empirical Methods in
  Natural Language Processing ({EMNLP})}, pages 1532--1543, Doha, Qatar, 2014.
  Association for Computational Linguistics.

\bibitem{mikolov2013distributed}
Tom{\'{a}}s Mikolov, Ilya Sutskever, Kai Chen, Gregory~S. Corrado, and Jeffrey
  Dean.
\newblock Distributed representations of words and phrases and their
  compositionality.
\newblock In Christopher J.~C. Burges, L{\'{e}}on Bottou, Zoubin Ghahramani,
  and Kilian~Q. Weinberger, editors, {\em Advances in Neural Information
  Processing Systems 26: 27th Annual Conference on Neural Information
  Processing Systems 2013. Proceedings of a meeting held December 5-8, 2013,
  Lake Tahoe, Nevada, United States}, pages 3111--3119, 2013.

\bibitem{kingma2014adam}
Diederik~P. Kingma and Jimmy Ba.
\newblock Adam: {A} method for stochastic optimization.
\newblock In Yoshua Bengio and Yann LeCun, editors, {\em 3rd International
  Conference on Learning Representations, {ICLR} 2015, San Diego, CA, USA, May
  7-9, 2015, Conference Track Proceedings}, 2015.

\bibitem{prakash2016neural}
Aaditya Prakash, Sadid~A. Hasan, Kathy Lee, Vivek Datla, Ashequl Qadir, Joey
  Liu, and Oladimeji Farri.
\newblock Neural paraphrase generation with stacked residual {LSTM} networks.
\newblock In {\em Proceedings of {COLING} 2016, the 26th International
  Conference on Computational Linguistics: Technical Papers}, pages 2923--2934,
  Osaka, Japan, 2016. The COLING 2016 Organizing Committee.

\bibitem{papineni2002bleu}
Kishore Papineni, Salim Roukos, Todd Ward, and Wei-Jing Zhu.
\newblock {B}leu: a method for automatic evaluation of machine translation.
\newblock In {\em Proceedings of the 40th Annual Meeting of the Association for
  Computational Linguistics}, pages 311--318, Philadelphia, Pennsylvania, USA,
  2002. Association for Computational Linguistics.

\bibitem{lin2004rouge}
Chin-Yew Lin.
\newblock {ROUGE}: A package for automatic evaluation of summaries.
\newblock In {\em Text Summarization Branches Out}, pages 74--81, Barcelona,
  Spain, 2004. Association for Computational Linguistics.

\bibitem{sun2012joint}
Hong Sun and Ming Zhou.
\newblock Joint learning of a dual {SMT} system for paraphrase generation.
\newblock In {\em Proceedings of the 50th Annual Meeting of the Association for
  Computational Linguistics (Volume 2: Short Papers)}, pages 38--42, Jeju
  Island, Korea, 2012. Association for Computational Linguistics.

\bibitem{madnani2012re}
Nitin Madnani, Joel Tetreault, and Martin Chodorow.
\newblock Re-examining machine translation metrics for paraphrase
  identification.
\newblock In {\em Proceedings of the 2012 Conference of the North {A}merican
  Chapter of the Association for Computational Linguistics: Human Language
  Technologies}, pages 182--190, Montr{\'e}al, Canada, 2012. Association for
  Computational Linguistics.

\bibitem{wubben2010paraphrase}
Sander Wubben, Antal van~den Bosch, and Emiel Krahmer.
\newblock Paraphrase generation as monolingual translation: Data and
  evaluation.
\newblock In {\em Proceedings of the 6th International Natural Language
  Generation Conference}, 2010.

\bibitem{chen2020semantically}
Wenqing Chen, Jidong Tian, Liqiang Xiao, Hao He, and Yaohui Jin.
\newblock A semantically consistent and syntactically variational
  encoder-decoder framework for paraphrase generation.
\newblock In {\em Proceedings of the 28th International Conference on
  Computational Linguistics}, pages 1186--1198, Barcelona, Spain (Online),
  2020. International Committee on Computational Linguistics.

\bibitem{zhangbertscore}
Tianyi Zhang, Varsha Kishore, Felix Wu, Kilian~Q Weinberger, and Yoav Artzi.
\newblock Bertscore: Evaluating text generation with bert.
\newblock In {\em International Conference on Learning Representations}.

\end{thebibliography}

\end{document}